\documentclass{article}

\usepackage{neurips_2022}

\usepackage{blindtext}
\usepackage{bbding}
\usepackage[utf8]{inputenc} 
\usepackage[T1]{fontenc}    
\usepackage[colorlinks = true,
            linkcolor = blue,
            urlcolor  = blue,
            citecolor = blue,
            anchorcolor = blue]{hyperref}
\usepackage{url}            
\usepackage{booktabs}       
\usepackage[dvipsnames]{xcolor}
\usepackage{amsfonts}       
\usepackage{nicefrac}       
\usepackage{microtype}      
\usepackage{xcolor}         
\usepackage{amsmath,mleftright,mathtools}
\usepackage{amssymb}
\usepackage{caption}
\usepackage{subcaption}
\usepackage{graphicx}
\usepackage{wrapfig}
\usepackage{float}
\usepackage{multirow}
\usepackage[toc,page]{appendix}
\usepackage{array}
\usepackage{diagbox}
\usepackage{algorithm}
\usepackage{algorithmic}
\usepackage{eucal}
\usepackage{caption} 
\def \methodname{{{Adaptive Gradient Variance Modulator}}}

\def \AGVM{{{AGVM}}}
\newcommand{\eg}{\emph{e.g.,}~}
\newcommand{\ie}{\emph{i.e.,}~}
\title{Large-batch Optimization for Dense Visual Predictions}

\author{%
  Zeyue Xue\footnotemark[1] \\
 The University of Hong Kong\\
 \texttt{xuezeyue@connect.hku.hk} \\
   \And
   Jianming Liang\footnotemark[1] \\
   Beihang University \\
   \texttt{ljmmm1997@gmail.com} \\
   \AND
   Guanglu Song \\
   Sensetime Research \\
   \texttt{songguanglu@sensetime.com} \\
   \And
   Zhuofan Zong\footnotemark[1] \\
   Beihang Univerisity \\
   \texttt{zongzhuofan@gmail.com} \\
   \And
   Liang Chen\footnotemark[1] \\
   Peking University \\
   \texttt{clandzyy@pku.edu.cn} \\
   \And
   Yu Liu\footnotemark[2] \\
   Sensetime Research \\
   \texttt{liuyuisanai@gmail.com} \\
   \And
   Ping Luo\footnotemark[2] \\
   The University of Hong Kong,\\ Shanghai AI Laboratory\\
   \texttt{pluo@cs.hku.hk} \\
}

\begin{document}

\maketitle

\renewcommand{\thefootnote}{\fnsymbol{footnote}}
\footnotetext[1]{Work done during an internship at Sensetime Research.}
\footnotetext[2]{Corresponding authors.}

\begin{abstract}
Training a large-scale deep neural network in a large-scale dataset is challenging and time-consuming. The recent breakthrough of large-batch optimization is a promising way to tackle this challenge. However, although the current advanced algorithms such as LARS and LAMB succeed in classification models, the complicated pipelines of dense visual predictions such as object detection and segmentation still suffer from the heavy performance drop in the large-batch training regime.
To address this challenge, we propose a simple yet effective algorithm, named \methodname{} (\AGVM{}), which can train dense visual predictors with very large batch size, enabling several benefits more appealing than prior arts.
Firstly, \AGVM{} can align the gradient variances between different modules in the dense visual predictors, such as backbone, feature pyramid network (FPN), detection, and segmentation heads. We show that training with a large batch size can fail with the gradient variances misaligned among them, which is a phenomenon primarily overlooked in previous work. Secondly, \AGVM{} is a plug-and-play module that generalizes well to many different architectures (\eg CNNs and Transformers) and different tasks (\eg object detection, instance segmentation, semantic segmentation, and panoptic segmentation). It is also compatible with different optimizers (\eg SGD and AdamW). Thirdly, a theoretical analysis of \AGVM{} is provided.
Extensive experiments on the COCO and ADE20K datasets demonstrate the superiority of \AGVM{}. For example, it can train Faster R-CNN+ResNet50 in 4 minutes without losing performance. \AGVM{} demonstrates more stable generalization performance than prior arts under extremely large batch size (\ie 10k). It enables training an object detector with one billion parameters in just 3.5 hours, reducing the training time by 20.9$\times$, whilst achieving 62.2 mAP on COCO. The deliverables are released at \url{https://github.com/Sense-X/AGVM}.

\end{abstract}

\section{Introduction}

The recent successes in many tasks of dense visual predictions rely on the large-scale datasets {\cite{deng2009imagenet, lin2014microsoft, Cordts2016Cityscapes}}, the increase of computational power (\eg GPUs), and the parallel training paradigm with large sample batches.
%
Sufficient computational resource enables large-batch training,  greatly reducing the training time \cite{you2017large}.
However, although simply scaling the batch size allows fewer iterations to update the parameters of deep neural networks, it often leads to dramatic drop of generalization performance  {\cite{goyal2017accurate, you2019large, keskar2016large}}.

To reduce the generalization gap in the  large-batch training paradigm, LARS {\cite{you2018imagenet}} scales the batch size of a plain ResNet50 from 8k to 32k without losing accuracy, enabling  to train an image classification model on ImageNet in a few minutes.
However, different from the plain network architectures in ImageNet classification \cite{chen2022geometric, chen2022mutual, chen2022evidential}, many tasks of dense visual predictions, such as object detection~\cite{girshick2015fast, Lin_2017_ICCV, tian2019fcos, carion2020end, song2020revisiting} and segmentation~\cite{he2017mask, bolya2019yolact, fang2021instances, xie2021segformer}, are solved by more complicated pipelines, which consist of multiple different modules, such as region proposal network (RPN)~\cite{girshick2015fast},  feature pyramid network (FPN)~\cite{lin2017feature}, detection head, and segmentation head.
Nevertheless, the recent advanced large-batch optimization methods such as LARS \cite{you2018imagenet} and LAMB \cite{you2019large}
are typically not sufficient to achieve good generalization performance in dense visual predictions.
The long training time of dense predictors greatly limits the researchers from making full use of the increasing computational power and large-scale datasets.

\begin{figure}[t] 
\centering 
\includegraphics[width=0.98\textwidth]{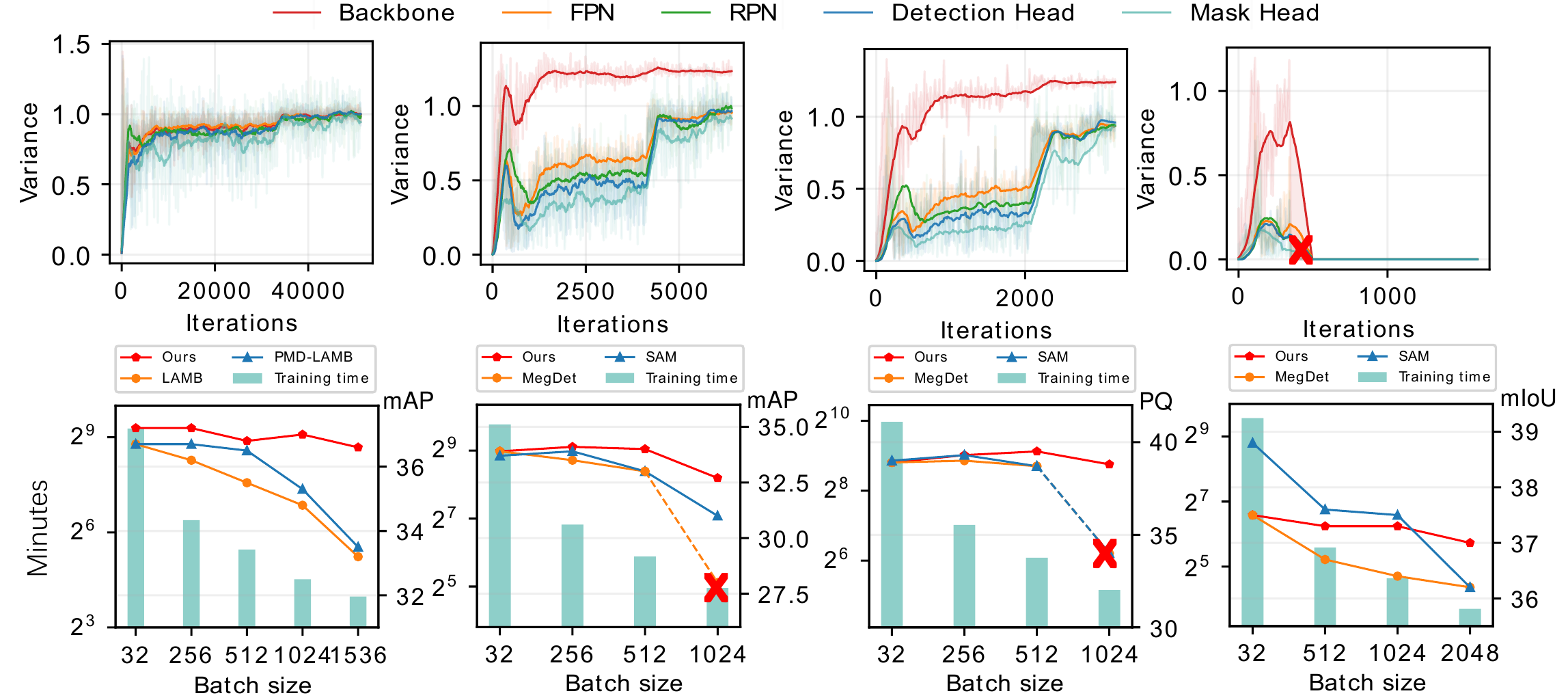} 
\caption{\small{
\textbf{First row}: 
Comparisons of the gradient variances (omitting learning rate in $\Phi_t^{(i)}$ referred to Eq.~\eqref{doc-modified}) of different network modules in Mask R-CNN, including backbone, FPN, RPN, and heads.
From left to right, the models are trained using SGD with a mini-batch size of 32, 256, 512, and 1024, respectively. 
Note that smaller batch size (32 in the first figure) produces similar $\Phi_t^{(i)}$
between different modules.
When the batch size increases from 256 to 1024 ($2^{\mathrm{nd}}\sim 4^{\mathrm{th}}$ figures), the gradient variance curves suffer from heavy misalignment between  modules. Specifically, the gradient variances are significantly small in the RPN, FPN, detection head, and mask head. We find that the larger the variance 
gap, the lower the model performance (the best performance is achieved when batch size equals 32). \textbf{Second row}: In figures from left to right, we compare the performance (right vertical axis) and training time of AGVM (bar diagram, left vertical axis) in different visual tasks, including object detection ($1^{\mathrm{st}}$ figure), instance segmentation ($2^{\mathrm{nd}}$), panoptic segmentation ($3^{\mathrm{rd}}$), and semantic segmentation ($4^{\mathrm{th}}$), where  the models are trained using different methods with different batch sizes.
The ``\textcolor{red}{$\times$}'' indicates training failure when using previous methods. Our method outperforms the recent approaches in all tasks with various batch sizes, significantly reducing training time.} }
\label{fig:motivation} 
\end{figure}


To address the above challenge, we present a novel large-batch training algorithm, named \methodname{} ({\AGVM{}}), which can train different complicated dense predictors with very large batch size, significantly reducing their training time while maintaining the generalization performance.
The design of \AGVM{} is motivated by a training phenomenon overlooked in prior arts. We call it gradient variance misalignment, which would present when a visual dense prediction pipeline contains many different modules and is trained with a large mini-batch, where different modules (\eg backbone, RPN, FPN, and heads) can have different gradient variance magnitudes, impeding the generalization ability.

As shown in the first row of Fig.\ref{fig:motivation}, where Mask R-CNN \cite{he2017mask} with ResNet50 \cite{he2016deep} as the backbone is trained using different batch sizes, we compare the gradient variances of different network modules, including backbone, FPN, RPN, detection head, and mask head. We see that when the batch size is small ($32$ in the first figure), the gradient variances of different network modules are similar throughout the training process. When the batch size increases from 256 to 1024 ($2^{\mathrm{nd}}\sim 4^{\mathrm{th}}$ figures), the gradient variances misalign in different modules whose variance gap enlarges during training. Training fails when batch size equals 1024.
More importantly,  the gradient variances have significantly smaller values 
in the RPN, FPN, detection head, and mask head compared to that in the backbone, and their gradient variances change  sharply in the late stage of training (two figures in the middle).
We find that  such misalignment undesirably burdens the large-batch training, leading to severe performance drop and even training failure. More observations on various visual tasks and networks can be found in Appendix \textcolor{blue}{A.2}.

The above empirical analysis naturally inspires us to design a simple yet effective method AGVM for training  dense visual predictors with multiple modules using very large batch size.
\AGVM{} directly modulates the misaligned variance of gradient,   making it consistent between different network modules throughout training. 
As shown in the second row of {Fig.\ref{fig:motivation}},
\AGVM{} significantly outperforms the recent approaches of large-batch training in four different visual prediction tasks with various batch sizes from 32 to 2048.
For example,
\AGVM{} enables us to train an object detector with a huge batch size 1536 (where prior arts may fail), reducing training time by more than 35$\times$ compared to the regular training setup.
  
This work makes three main \textbf{contributions}. \textbf{Firstly}, we carefully design \AGVM{}, which to our knowledge, is the first large-batch optimization method for various dense prediction networks and tasks. We evaluate \AGVM{}  in different  architectures (\eg CNNs and Transformers), solvers (\eg SGD and AdamW), and tasks (\eg object detection, instance
segmentation, semantic segmentation, and panoptic segmentation).
\textbf{Secondly}, we provide a convergence analysis of \AGVM{}, which converges to a stable point in a general non-convex optimization setting. We also conduct an empirical analysis that reveals an important insight: the inconsistency of {{effective batch size}} between different modules would aggravate the gradient variance misalignment when batch size is large, leading to performance drop and even training failure. We believe this insight may facilitate future research for large-scale training of complicated vision systems. 
\textbf{Thirdly}, extensive experiments are conducted to evaluate \AGVM{}, which achieves many new state-of-the-art performances on large-batch training. For example, 
 \AGVM{} demonstrates more stable generalization performance than prior arts under extremely large batch size (\ie 10k).
In particular, it enables training of the widely-used Faster R-CNN+ResNet50 within 4 minutes without performance drop.
More importantly, \AGVM{} can train a detector with one billion parameters within just 3.5 hours, which reduces the training time by 20.9$\times$, while achieving a top-ranking mAP 62.2 on the COCO dataset.

\section{Preliminary and Notation}
Let $S=\left\{\left(x_{i},y_{i}\right)\right\}_{i=1}^{n}$ denote a dataset with $n$ training samples, where $x_i$ and $y_i$ represent a data point and its label respectively. 
We can estimate the value of a loss function $L:\mathbb{R}^{d} \rightarrow \mathbb{R}$ using a  mini-batch of samples that  are randomly sampled, and obtain $l(w_{t})= \frac{1}{b} \sum_{j \in S_{t}} L\left(w_{t},(x_{j},y_{j})\right)$, where $S_{t}$ denotes the mini-batch at the $t$-th iteration  with batch size $\left|S_{t}\right|=b$ and $w_{t}$ represents the parameters of a deep neural network. 
We can apply stochastic gradient descent (SGD), one of the most representative algorithms, to update the parameters $w_{t}$. The SGD update equation with learning rate $\eta_{t}$ is:
\begin{equation}
    w_{t+1}=w_{t}-\eta_{t}\nabla l(w_{t}),
\end{equation}
where $\nabla l(w_{t})$ represents the gradient of the loss function with respect to $w_{t}$.

\textbf{Layerwise Scaling Ratio.} 
In large-batch training, \citet{you2018imagenet} observe that the ratio between the norm of the layer weights and the norm of the gradients is unstable (\ie oscillate a lot),  leading to training failure. {\citet{you2018imagenet}} present the LARS algorithm, which adopts a layerwise scaling ratio, $\|w^{(i)}_{t}\|/\|\nabla l(w^{(i)}_{t})+\lambda w^{(i)}_{t}\|$, to modify the magnitude of the gradient of the $i$-th layer  $\nabla l(w^{(i)}_{t})$, where $w^{(i)}_{t}$ and $\lambda$ indicate the parameters of the $i$-th layer and the  weight decay coefficient, respectively.
Furthermore, LAMB {\cite{you2019large}} improves LARS by combining the AdamW optimizer with the  layerwise scaling ratio. 
%
It can be formulated as $r_{t}=m_{t}/\sqrt{v_{t}+\epsilon}$, where $m_{t} = \beta_{1}m_{t-1}+(1-\beta_{1})\nabla l(w_{t})$ and $v_{t} = \beta_{2}v_{t-1}+(1-\beta_{2})\nabla l(w_{t})^2$. 
The layerwise scaling ratio of LAMB can be computed by $\|w^{(i)}_{t}\|/\|r_{t}^{(i)}+\lambda w^{(i)}_{t}\|$.

\textbf{Sharpness-aware Minimization.} 
Large-batch training often converges to a sharp local minima, resulting in undesired generalization performance.
The sharpness-aware minimization (SAM) {\cite{foret2020sharpness}} algorithm explicitly penalizes the sharp minima and finds the parameters whose neighbors (in an $l_{p}$-ball) have low training loss function values using the following objective function:
\begin{equation}
l^{\mathrm{SAM}}(w_{t})=\max _{\|\epsilon\|_{p} \leq \rho} l(w_{t}+\epsilon).
\label{eq:SAM_1}
\end{equation}
To solve the above equation, SAM applies one-step gradient ascent to determine $\epsilon=\rho \nabla l(w_{t})/\|\nabla l(w_{t})\|$.
Its gradient is then approximated by $\nabla l^{SAM}(w_{t})\approx \nabla l(w_{t})|_{w_{t}+\epsilon}$.
However,  SAM involves two sequential gradient computations at each iteration and thus doubles the computational cost.

\textbf{Gradient Variance Estimation.}
\citet{NEURIPS2021_abea47ba} utilize the cosine similarity between two aggregated gradients from the replicas in a distributed training system, to estimate the gradient variance {between SGD and GD} efficiently.
Specifically, we can compute the gradient for each  sample in the $t$-th mini-batch $S_t$ of batch size $b$, denoted by $r_{1,t},..., r_{j,t},...,r_{b,t}$. We have $\nabla l(w_{t}) = \frac{1}{b}\sum^b_{j=1}r_{j,t}$.
We split the above gradients into two groups
and average each group, obtaining $G_{t,1}=\frac{2}{b} \sum_{j=1}^{\frac{b}{2}} r_{2 j-1,t}$ and $G_{t,2}=\frac{2}{b} \sum_{j=1}^{\frac{b}{2}} r_{2 j,t}$, respectively. Then the gradient variance can be measured  by $\Phi_t =  1-\cos(G_{t,1},G_{t,2})$, where $\cos(\cdot,\cdot)$ is the cosine similarity function.

\section{Our Approach}

Our goal is to perform large-batch training for dense visual predictors with many different network modules.
As illustrated in Fig.\ref{fig:motivation}, the inconsistency of gradient variances among different modules need to be modulated.


\textbf{Gradient Variance across Modules.} 
We derive an
updated (considering learning rate)
gradient variance to delve into the difference of network modules in complicated dense visual prediction pipelines. The updated gradient variance of the  $i$-th network module at the $t$-th iteration can be formulated as:
\begin{equation}
\mathrm{Var}(\eta_{t} g_t^{(i)})= \frac{n-b}{2n-b}\underbrace{  \eta_{t}^{2}(1-\mathbb{E}[\cos(G_{t,1}^{(i)},G_{t,2}^{(i)})])}_{\Phi^{(i)}_{t}}\mathbb{E}[\| g_t^{(i)}\|^{2}],
\label{doc-modified}
\end{equation}
where $n$ and $b$ are the number of training samples and the mini-batch size, respectively. $\eta_{t}$ is the learning rate. 
$g_{t}^{(i)}$ indicates the gradient of the $i$-th network module. $G_{t,1}^{(i)}$ and $G_{t,2}^{(i)}$ are two groups of the gradient estimation as discussed above.
Since each entry in the vector $g_t^{(i)}$ could be assumed \textit{i.i.d.} in a massive dataset following {\cite{NEURIPS2021_abea47ba,pmlr-v119-wu20c}},  $\Phi_t^{(i)}$ is thus proportional to the above updated gradient variance. At each training iteration, we can approximate the updated gradient variance by
$\Phi_t^{(i)} =  \eta_{t}^{2}(1-\cos(G_{t,1}^{(i)},G_{t,2}^{(i)}))$.
Note that $\Phi_t^{(i)}$ for $i$-th module has been normalized by the number of parameters, so $\Phi_t^{(i)}$ of different modules are comparable.
For consistency of presentation, we still call $\Phi_t^{(i)}$ gradient variance, which enables us to estimate the gradient variance of each network module at each training iteration.
More discussions can be found in Appendix \textcolor{blue}{A.1}.

\textbf{Adaptive Gradient Variance Modulator (\AGVM{}).}
Let $\mathcal{M}$ be a set of modules in a complicated dense prediction pipeline, where $\mathcal{M}$ has  $h$ different modules. At the $t$-th iteration, we have a set of learning rates, $\{\hat{\eta}^{(i)}_t | i\in\{1,2,...,h\}\}$,  corresponding to different modules. We treat  the $\texttt{Backbone}$ ($i=1$)  as the anchor and modulate other modules making their gradient variances consistent with the $\texttt{Backbone}$.
Specifically, we adjust the module learning rates $\hat{\eta}_t^{(i)}$ by using
the ratio between $\Phi^{(1)}_t$ and $\Phi^{(i)}_t$.
The update rule for each network module can be written as:
\begin{equation}
w^{(i)}_{t+1} = w^{(i)}_{t}-\hat{\eta}_t^{(i)} g^{(i)}_{t},~~\mathrm{where~~} \hat{\eta}_t^{(i)} = \eta_{t} \mu^{(i)}_t ~~\mathrm{and~~} \mu^{(i)}_t=\sqrt{\frac{\Phi^{(1)}_t }{\Phi^{(i)}_t}},
\label{sgd-gvc}
\end{equation}
where $ \eta_{t}$ is the global learning rate. 
However, simply adjusting the learning rates on-the-fly would easily yield training failure due to the transitory large variance ratio that impedes the optimization. We propose a momentum update to address this problem.
Let $\alpha\in[0,1)$ be a momentum coefficient, we have:
\begin{equation}
\mu_t^{(i)} \gets \alpha \mu_{t-1}^{(i)} + (1-\alpha)\mu_t^{(i)},
\label{moving}
\end{equation}
which can reduce the influence of unstable variance. Note that we update $\mu_t^{(i)}$ each $\tau$ iterations.

\textbf{Discussion on Momentum and Weight Decay.}
In practice, the weight decay is widely used as a regularizer and is tightly coupled with the learning rate and the momentum.
For instance, the gradient $g_t^{(i)}$ will be replaced by the momentum, such as $m^{(i)}_{t} = \beta_{1} m^{(i)}_{t-1} + (1 - \beta_{1}) (g^{(i)}_{t}+\lambda w^{(i)}_t)$ ~{\cite{you2019large,smith2018disciplined}}, where $\beta_1$ and $\lambda$ indicate the momentum coefficient and the weight decay coefficient, respectively.
We observe that it's also important to modulate the learning rate by Eq.\eqref{sgd-gvc} when  weight decay is presented.
In addition, 
since the above $m_t$ is a momentum-based moving average of $(g^{(i)}_{t}+\lambda w^{(i)}_t)$,
we can directly apply $\hat{\eta}_t^{(i)}$ onto $m_t^{(i)}$.

\textbf{Extensions to Different Optimization Algorithms.} 
\AGVM{} can be easily embedded into different optimization algorithms such as SGD and AdamW. We demonstrate the details in Appendix \textcolor{blue}{A.6}: Alg.1 and Alg.2, respectively.
They can be easily implemented using a deep learning framework \eg PyTorch~\cite{paszke2019pytorch}.

\textbf{Discussion on Convergence Rate.} With AGVM, the SGD and the AdamW optimizers still have appealing convergence properties in the general non-convex settings. Considering some mild assumptions in stochastic optimization and the case without heavy-ball momentum ($\beta_{1}=0$), SGD and AdamW achieve $O(1/\sqrt{T})$ and $O(\ln(T)/\sqrt{T})$ convergence rate respectively with appropriate choice of the learning rate $\eta_{t}$. We present the analysis in Appendix \textcolor{blue}{A.4}.

\begin{table}[t]
\centering
\caption{\small{\textbf{Comparisons between different methods.} ``Generalization'' indicates the methods' generalization ability for dense visual prediction tasks. The number of ``+'' in the column ``stable to batch size scaling'' means the degree of stability when batch size is increased, whereas the number in the bracket means the maximum applicable batch size without divergence on object detection. We measure the average extra overhead of the Faster R-CNN+ResNet50 detector at each iteration using 128 NVIDIA A100 GPUs (total batch size is 1024). The number in the column ``extra overhead'' indicates the ratio of extra overhead (an extra all-reduce call) compared to the original computations. ``N/A'' means no extra overhead.}}
\label{table:comparison}
\scalebox{0.77}{
\begin{tabular}{c|p{130pt}<{\centering}|p{60pt}<{\centering}|p{80pt}<{\centering}|p{60pt}<{\centering}|p{40pt}<{\centering}}
\hline
  Method &Solution  & Generalization  & Less hyperparam. tuning & Stable to batch size scaling  &   Extra overhead     \\ \hline
  MegDet \cite{peng2018megdet}&Accumulate statistics of BN&\CheckmarkBold   & \CheckmarkBold   & + (1024)  &  N/A  \\ \hline
  SAM \cite{foret2020sharpness}& Penalize  sharp minima & \XSolidBrush   &  \XSolidBrush  & + (2048) & 100\%  \\ \hline
  LARS \cite{you2018imagenet}& Rectify layerwise gradient &  \XSolidBrush  &  \XSolidBrush  & + (1024)  &   N/A \\ \hline
  LAMB \cite{you2019large}& Rectify layerwise gradient &  \XSolidBrush   & \XSolidBrush  & ++ (4096)  &  N/A  \\ \hline
 PMD-LAMB \cite{wang2020large}& Reduce historical effect& \CheckmarkBold  & \XSolidBrush  & ++ (4096)  &  N/A  \\ \hline
  AGVM &Balance gradient variance& \CheckmarkBold  &  \CheckmarkBold  & +++ (10k) &  0.12\% \\ \hline
\end{tabular}
}
\end{table}

\textbf{Comparisons with Existing Works.}
The purpose of exploring large-batch training is to speed up model training with increasing computational power, as well as enabling us to explore the larger dataset.
As shown in Table~\ref{table:comparison},
the seminal works such as LARS{~\cite{you2018imagenet}}, LAMB{~\cite{you2019large}}, and SAM{~\cite{foret2020sharpness}} have made great contributions to large-batch training for plain vision pipelines \eg image-level prediction,  despite that they often require hyper-parameter tuning by experienced engineers.
For complicated pipelines of dense visual predictions, they are typically not sufficient to achieve desired generalization performance. 
MegDet {\cite{peng2018megdet}} and PMD-LAMB {\cite{wang2020large}} contribute the preliminary attempts by applying large-batch training on object detection.
Different from these approaches, 
we revisit the design paradigm of the complicated dense visual perception pipelines and present a simple yet effective solution, \AGVM{}, which is insensitive to hyperparameter tuning and can be easily plugged into many visual perception pipelines.
For example, \AGVM{} can perform stable training with an unprecedented batch size 10K, which could greatly reduce the training time.
Moreover, \AGVM{} adds a negligible computational overhead in training, unlike SAM which involves two sequential (non-parallelizable) gradient computations at each iteration, resulting in a significant increase of the training time.

\section{Experiments} \label{experiments}

\textbf{Dataset.}  We conduct comprehensive experiments on the MS-COCO 2017{~\cite{lin2014microsoft}} and the ADE20K{~\cite{zhou2017scene}} datasets. 
Specifically, we perform various tasks of object detection, instance segmentation, and panoptic segmentation  on COCO, and conduct semantic segmentation on ADE20K. 

\textbf{Baselines.}
Since the prior arts of {large-batch} optimization methods can be divided into two types, SGD-based methods (\ie LARS \cite{you2018imagenet}, MegDet~\cite{peng2018megdet}) and AdamW-based methods (\ie LAMB \cite{you2019large}, PMD-LAMB \cite{wang2020large}).
For fair comparison, we introduce two  training configurations using SGD and AdamW with AGVM, respectively. 
The details of the hyper-parameter settings can be found in Appendix \textcolor{blue}{A.5}.

\textbf{Pipelines and Models.}
To evaluate the generalization ability of AGVM, we conduct extensive experiments on different pipelines, including RetinaNet {\cite{lin2017focal}}, Faster R-CNN {\cite{girshick2015fast}}, Mask R-CNN {\cite{he2017mask}}, Panoptic FPN {\cite{kirillov2019panoptic}}, and Semantic FPN {\cite{kirillov2019panoptic}}. For the backbone networks, we use ResNet {\cite{he2016deep}} and Swin Transformer {\cite{liu2021swin}}. We strictly follow the official implementations of these pipelines and models.

\textbf{Implementation Details.}
We implement \AGVM{} in PyTorch and reproduce PMD-LAMB with the official implementation of LAMB {\cite{you2019large}}. We also evaluate  LARS {\cite{you2018imagenet}} and SAM {\cite{foret2020sharpness}} by borrowing their official implementations. To make fair comparisons, we follow the same learning rate scaling method in all experiments. For SGD optimizer, we use linear learning rate scaling when batch size is less than 128 (256 on semantic segmentation). When the batch size is greater than 128, we use the square root of learning rate scaling to avoid divergence in the training process. For PMD-LAMB and LAMB, we follow the learning rate scaling scheme in \cite{wang2020large}. 
We apply a learning rate warm-up scheme to avoid divergence when the  learning rate is large. The implementation details can be found in Appendix \textcolor{blue}{A.5}.

\begin{table*}[t]
\small
\centering
\caption{\small{ \textbf{Comparisons} in different tasks (\ie object detection, instance segmentation,
semantic segmentation, and panoptic segmentation) and pipelines (\ie Faster R-CNN, Mask R-CNN, Semantic FPN, and Panoptic FPN). All pipelines use ResNet50 as the backbone and we use SGD as optimizer. We see that previous methods' performances drop a lot when scaling the batch size and even result in training failure when batch size is 1024 (``NaN''). 
Since LARS always leads to huge performance drop in large-batch settings, so we only report its performance on Mask R-CNN. We also report the comparisons with MegDet and SAM. The best-performing models are shown in bold. Surprisingly, \AGVM{} can alleviate the training difficulties in large-batch settings.
}}
\label{tbl:sgd_summary}
\begin{center}
\scalebox{0.93}{
\begin{tabular}{
ccccccccc
}
\toprule
\multirow{2}{*}{Pipeline} &
\multirow{2}{*}{Dataset} &
\multirow{2}{*}{Task} &
\multirow{2}{*}{Batch size}
& \multicolumn{4}{c}{Performance}
& \multirow{2}{*}{Iterations} \\
\cmidrule(lr){5-8}
& & & &
MegDet & SAM & LARS & Ours &
\\
\midrule
\multirow{4}{*}{Faster R-CNN} &\multirow{4}{*}{COCO} &\multirow{4}{*}{Detection} & 32 & 36.8  & 36.0 & - & \textbf{36.8} & 58640 \\
  & & & 256 & 36.1 & 36.5 & - &\textbf{36.7} & 7344  \\
  & & & 512 & 35.8 & 35.7 & - &\textbf{36.7} & 3680  \\
  & & & 1024 & 34.2 & 33.0 & - &\textbf{35.4} & 1840  \\
\midrule
\multirow{4}{*}{Mask R-CNN} &\multirow{4}{*}{COCO} &\multirow{4}{*}{Instance Seg} & 32 & 33.9 & 33.7 & \textbf{34.0}  &33.9  & 51310  \\
& & & 256 & 33.7 & 33.9 &32.0& \textbf{34.1} & 6426 \\
& & & 512 & 33.1 & 33.0 & 30.4&\textbf{33.9} & 3220  \\
&& & 1024 & NaN & 31.0 &25.1& \textbf{32.6} & 1610  \\
\midrule
\multirow{4}{*}{Semantic FPN} &\multirow{4}{*}{ADE20K} &\multirow{4}{*}{Semantic Seg} & 32 & 37.5 & \textbf{38.8}  &-& 37.5 & 160000  \\
 & & & 512 & 36.7 & \textbf{37.6} &- & 37.3 & 10000  \\
 & & & 1024 & 36.4 & \textbf{37.5} & - &37.3 & 5000\\
 & & & 2048 & 36.2 & 36.2 & - &\textbf{37.0} & 2500 \\
\midrule
\multirow{4}{*}{Panoptic FPN} &\multirow{4}{*}{COCO} &\multirow{4}{*}{Panoptic Seg} & 32 & 38.9& \textbf{39.0} & -  & 38.9 & 51310  \\
&  & & 256 & 39.2& 39.3 &- & \textbf{39.3} & 6426 \\
& &  & 512 &  38.7 & 38.7 & - &\textbf{39.5} & 3220 \\
&  & & 1024 & NaN & NaN &- & \textbf{38.8} & 1610\\
\bottomrule
\end{tabular}}
\end{center}
\vspace{-5mm}
\end{table*}

\begin{table*}[t]
\small
\caption{\small{\textbf{Comparisons} of performance for object detection on the COCO dataset with different backbones and batch sizes. We compare the mAP and the number of iterations of AdamW, LAMB, PMD-LAMB, and \AGVM{}+AdamW. The best-performing models are shown in bold. The underlined numbers indicate the results are borrowed directly from 
{\cite{wang2020large}.}
}}
\label{adamw}
\small{
\begin{center}
\scalebox{0.93}{
\begin{tabular}{
cccccccc
}
\toprule
\multirow{2}{*}{Pipeline} &
\multirow{2}{*}{Backbone} &
\multirow{2}{*}{Batch size}
& \multicolumn{4}{c}{Performance}
& \multirow{2}{*}{Iterations} \\
\cmidrule(lr){4-7}
& & & AdamW &
LAMB & PMD-LAMB& \AGVM{} (ours) &
\\
\midrule
\multirow{5}{*}{Faster R-CNN} &\multirow{5}{*}{ResNet50} & 32 & 37.1 & 36.7  & 36.7  & \textbf{37.1} & 43980 \\
 & & 256 & 36.9 & \underline{36.2} & \underline{36.7} &\textbf{37.2} & 5508  \\
 & & 512 &  36.2 & \underline{35.5} & \underline{36.5}  &\textbf{36.8} & 2760  \\
 & & 1024 & 36.2 & \underline{34.8} & \underline{35.3}  &\textbf{37.0} & 1380  \\
  & & 1536 & 35.9 &33.2 & 33.5  &\textbf{36.6} & 924  \\
\midrule
\multirow{4}{*}{Faster R-CNN} & \multirow{4}{*}{Swin-Tiny} & 32 & 43.6 & 42.9 & 40.2  & \textbf{43.7} & 47645  \\
& & 256& 43.4 & 43.5  &42.4& \textbf{43.5} & 5967 \\
& & 512& 42.7 & 42.9  &41.3&\textbf{43.2} & 2990  \\
& & 1024 & 42.4 & 41.6  &39.4& \textbf{42.8} & 1495  \\
\bottomrule
\end{tabular}
}
\end{center}
}
\vspace{-5mm}
\end{table*}

\subsection{Comparisons to the State-of-The-Art Methods}

Table \ref{adamw} compares the results of object detection on the COCO dataset with different backbones and batch sizes.  We compare the mAP and the number of iterations of LAMB, PMD-LAMB, and AGVM using the AdamW optimizer. 
To our knowledge, AGVM reports the first  result that successfully scales the batch size to 1536 with negligible performance drop compared to small-batch training using LAMB. 
We also see that \AGVM{} contributes significant improvements along with the continuous increase of the batch size.
By scaling the batch size larger than 1024 for different backbones, \AGVM{} can still achieve {36.6} and {42.8} mAP without heavy hyper-parameter tuning. In conclusion, compared with LAMB and PMD-LAMB, \AGVM{} achieves more accurate results whilst reducing training iterations and runtime. \AGVM{} can be embedded in CNN and Transformer models.

\textbf{Generalize to various pipelines, architectures, and optimizers.}
\AGVM{} can be generalized  to different tasks, pipelines, architectures, and optimizers. Table~\ref{tbl:sgd_summary} compares MegDet, SAM, LARS, and \AGVM{} in different dense visual prediction tasks, including object detection, instance segmentation, semantic segmentation, and panoptic segmentation on COCO and ADE20K. We evaluate four representative  pipelines (\eg Faster R-CNN, Mask R-CNN, Semantic FPN, and Panoptic FPN) with different batch sizes from 32 to 1024. We see that scaling the batch size only allows fewer iterations to update weights in previous methods, whose performances drop a lot and even have training failure  when the batch size is 1024 (denoted by ``NaN''). 
In contrast, \AGVM{} yields surprising results in all tasks when increasing the batch size.
Table~\ref{optimizer} reports the performances of \AGVM{} trained with different optimizers, SGD and AdamW. \AGVM{} works well with both of them. 



\begin{minipage}[t]{0.48\textwidth}
\makeatletter\def\@captype{table}
\centering
\caption{\small{Training time of Faster R-CNN with batch size 2 per NVIDIA A100. }}
\label{table:faster time}
\scalebox{0.85}{
\begin{tabular}{cccccc}
\toprule
Batch size &  32  & 256  & 512 & 1024 & 1536 \\ 
\midrule
GPUs  &  16  & 128  & 256 & 512  & 768  \\ 
\midrule
Time (min)&   148 & 20.8 & 11.8 & 6.0    & \textbf{4.2}  \\ \bottomrule
\end{tabular}}
\end{minipage}
\hspace{2mm}
\begin{minipage}[t]{0.48\textwidth}
\makeatletter\def\@captype{table}
\centering
\caption{\small{Scaling the batch size to 10k on RetinaNet with ResNet18.}}
\label{table:10K}
\scalebox{0.85}{
\begin{tabular}{p{60pt}<{\centering}p{25pt}<{\centering}p{25pt}<{\centering} p{25pt}<{\centering}}
\toprule
Batch size & 32   & 4k  & 10k   \\
\midrule
PMD-LAMB & 31.4 & 23.5 & NaN \\
\midrule
Ours & 32.8 & \textbf{28.7} & \textbf{26.7}  \\\bottomrule
\end{tabular}}
\end{minipage}


\begin{minipage}[t]{0.48\textwidth}
\makeatletter\def\@captype{table}
\centering
\small
\caption{\small{\AGVM{}+different optimizers on Faster R-CNN. \AGVM{} works well with both these optimizers.}}
\label{optimizer}
\scalebox{0.85}{
\begin{tabular}{ccccc}
    \toprule
   Optimizer   & AGVM& Batch size & Backbone    & mAP  \\
    \midrule
   SGD   &\XSolidBrush &512 & ResNet50 &   35.8 \\
    SGD &\CheckmarkBold & 512 & ResNet50 & \textbf{36.7}\\
    \midrule
   AdamW  & \XSolidBrush&  512    & ResNet50 &   36.2 \\
    AdamW  &\CheckmarkBold &   512    & ResNet50 &  \textbf{36.8} \\
    \midrule
    AdamW  & \XSolidBrush&   512   & Swin-Tiny &   42.7 \\
    AdamW  &\CheckmarkBold &   512    & Swin-Tiny &   \textbf{43.2}\\
    \bottomrule
\end{tabular}}
\end{minipage}
\hspace{2.5mm}
\begin{minipage}[t]{0.48\textwidth}
\small
\makeatletter\def\@captype{table}
\centering
\caption{\small{Anchor module selection. We report the segmentation mAP with different anchor modules.}}
\label{table:align}
\scalebox{1.0}{
\begin{tabular}{ccc}
    \toprule
Pipeline  & Modules    & mAP  \\
    \midrule
   &Backbone   &  \textbf{33.9} \\
   & FPN & 33.3\\
 Mask R-CNN  & Detection Head  &  33.1 \\
    & RPN &  33.1 \\
   & Mask Head  &    32.9 \\
    \bottomrule
\end{tabular}}
\end{minipage}

\textbf{Training COCO in 4 minutes.}
With AGVM, we can push the frontier of fast training time on COCO.
%
We employ Faster R-CNN with ResNet50-FPN as the detector and use the same experimental setting as {\cite{wang2020large}}. Then we explore how fast AGVM can reach the 36.6 mAP@0.5:0.95 reported in {\cite{wang2020large}} (which needs 12 minutes to train). Different from the hardware setup in Fig.~\ref{fig:motivation} (batch size 8 per GPU), this experiment is conducted on 768 NVIDIA A100 GPUs.  As shown in Table \ref{table:faster time}, we reduce the original small-batch training time from 2.5 hours to only 4.2 minutes, which is the fastest record to our knowledge.

\textbf{Scaling the batch size to 10k.}
We also try to push the frontier of large batch size in dense visual prediction tasks. 
We choose RetinaNet with ResNet18 as the detector, which is trained for 24 epochs (2$\times$) using the AdamW optimizer.
For batch size 4k and 10k, the learning rates are 0.001 and 0.0015, respectively.
The mAP results on COCO are shown in Table~\ref{table:10K}.
Without bells and whistles, the batch size is successfully scaled to 10k while maintaining generalization ability, but PMD-LAMB fails (``NaN'').

\begin{table}[t]
\small
\vspace{-3mm}
\caption{\small{\textbf{Extending  UniNet \cite{liu2021uninet} to one billion parameters}. Both AdamW and PMD-LAMB do not converge when the batch size is 960. On the contrary, our method achieves a top-ranking mAP 62.2 on the COCO dataset, while reducing the training time by 20.9$\times$.}}
\label{table:1b}
    \centering
\scalebox{1.0}{
\begin{tabular}{cccccc}
    \toprule
   Optimizer   & Batch size     & Box mAP &Seg mAP & Iterations &Wall-clock time \\
    \midrule
   AdamW   & 32 & 62.6 & 53.8 & 43980 & 73 hours\\
    AdamW & 960 & NaN & NaN & - &- \\
    PMD-LAMB & 960 & NaN & NaN & -&- \\
   Ours  &   960    & 62.2 & 53.4 & \textbf{1349} & \textbf{3.5 hours}\\
    \bottomrule
\end{tabular}}
\vspace{-5mm}
\end{table}

\textbf{Scaling the detector to 1-Billion parameters.} We evaluate \AGVM{} on an extremely-large detector using the  UniNet \cite{liu2021uninet}. We extend it to one billion parameters by following the  design in {\cite{liu2021uninet}}. The detailed settings are released in Appendix \textcolor{blue}{A.5}.
Table~\ref{table:1b} shows that AGVM still stabilizes and accelerates the training process in such a large model regime. Both AdamW and PMD-LAMB diverge in the early training stage.
AGVM  can reduce the training time from 3 days (batch size 32) to 3.5 hours using 480 NVIDIA A100 GPUs, achieving a 62.2 box mAP on COCO test-dev benchmark, whilst reducing the training wall-clock time by more than 20 times. 

\subsection{Ablation Study}
\textbf{Insensitive to hyper-parameter $\tau$ and $\alpha$.} 
We study the effect of the interval parameter $\tau$, which means we update $\mu_{t}^{(i)}$ every $\tau$ iterations, as well as the coefficient of moving average $\alpha$ using Mask R-CNN. 
The experimental results in Table~\ref{update frequency} indicate that \AGVM{} is not sensitive to these two hyper-parameters. In practice, we employ $\tau=10$ and $\alpha=0.97$ by default. 
When the batch size is significantly large (\eg larger than 1K), we reduce the interval to $\tau=5$ to update $\mu_{t}^{(i)}$ faster.

\textbf{Anchor module selection.} In AGVM, we choose the {backbone} network as the anchor and modulate other modules to make their gradient variances consistent with the {backbone}. 
To deeply investigate this selection, we choose different modules as the anchors. As shown in Table~\ref{table:align}, we see that the {backbone} is the optimal anchor because the backbone network 
plays the most important role in dense visual predictions. 

\begin{table}[t]
\centering
\small
\caption{\small{
\textbf{Insensitive to hyper-parameter $\tau$ and $\alpha$.} We gradually decrease the update frequency of $\mu_{t}^{(i)}$ from left to right and report the Detection mAP and Segmentation mAP of Mask R-CNN. These results indicate \AGVM{} is not sensitive to these two hyper-parameters. However, when we don't introduce moving average coefficient, the training fails in the early stage.}}
\label{update frequency}
\centering
\begin{small}
\begin{tabular}{c|c|c|c|c|c|c}
\hline
$\tau$ / $\alpha$ & None & 5 / 0.95    & 5 / 0.97    & 10 / 0.97   & 20 / 0.97   & 20 / 0.98   \\ \hline
mAP             & NaN  & 37.5 / 33.9 & 37.5 / 34.0 & 37.5 / 33.9 & 37.6 / 33.9 & 37.5 / 34.0 \\ \hline
\end{tabular}
\end{small}
\vspace{-2mm}
\end{table}


\begin{figure}[t] 
\centering
\includegraphics[width=0.95\textwidth]{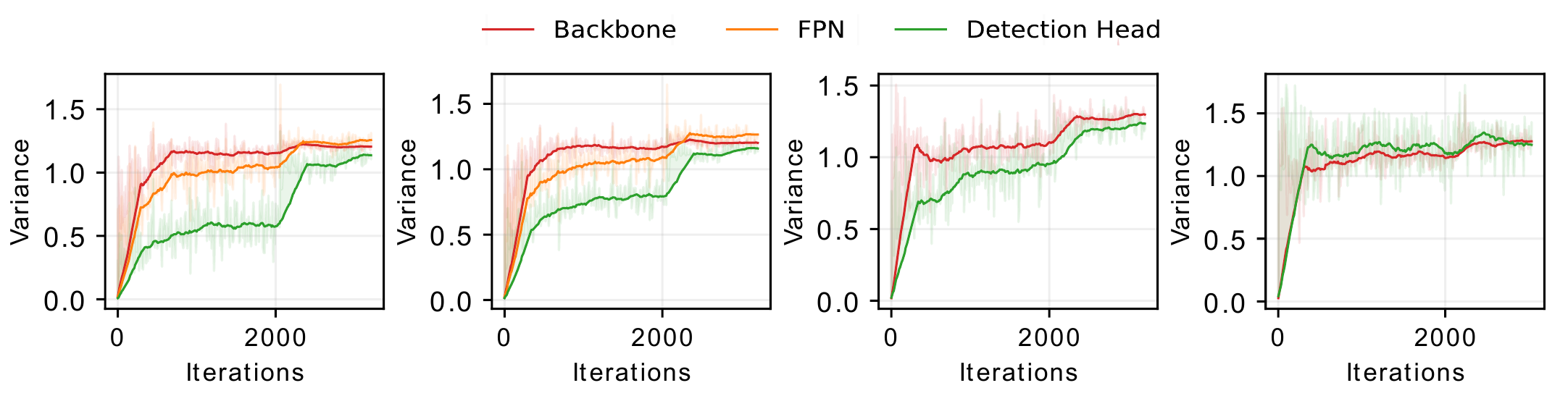}
\caption{\small{
\textbf{Ablative experiments on exploring the gradient variance misalignment.} To validate our result on effective batch size, we progressively use independent detection heads, remove FPN, and mask 75\% pixels to reduce the effective batch size on the detection head. Finally, we find a near-constant trend of variance throughout training towards convergence between the backbone and the detection head.}} 
\label{fig:analysis} 
\end{figure}

\textbf{Delving into the gradient variance misalignment.}
We answer an important question:
\textit{what causes the gradient variance misalignment for dense visual predictors?} To tackle this question, we revisit the data flow of dense prediction pipelines and find that
the {effective batch size} is not consistent between different network modules. For instance, due to the shared detection head (\ie classifiers and regressors) in all the levels of the FPN and different region proposals, the detection head has a different {effective batch size} compared to the backbone. Similarly, the RPN (or detection head in RetinaNet) shared by all FPN levels and pixel-wise loss computation lead to the increased {effective batch size} in RPN. Similar to a previous work \cite{pmlr-v119-wu20c}, we find that a larger effective batch size leads to lower gradient variance of modules (\eg RPN, detection head).

 
To explore these analyses, we conduct a progressive ablation study using the RetinaNet, 
as shown by the different gradient variance curves in Fig.\ref{fig:analysis}.
We have three observations.
(1) Intuitively, the shared head leads to the unavoidable batch size misalignment between the backbone and the detection head. For example, given an input mini-batch size ${B}$, the valid mini-batch size for the detection head is $N{B}$, where $N$ is the pyramidal feature number.
This motivates us to directly replace the shared detection head by independent detection heads. As illustrated by the second figure in Fig.\ref{fig:analysis}, the gradient variance misalignment between the detection head and the backbone has been significantly reduced.
(2) Furthermore, compared with the plain network architecture, we argue that the effective batch size is also related to the bottom-up and top-down pathways in FPN. To evaluate this, 
we remove FPN and only adopt the final-level feature map to perform detection. As shown in the third figure in Fig.\ref{fig:analysis}, this alleviates the variance difference between the backbone and the detection head. (3)
In the fourth figure, we randomly ignore 75\% pixels for loss computation in the predictions generated by detection head.
This  leads to a near-constant trend of variance throughout training towards convergence between the backbone and the detection head.
We have done a similar study using Faster R-CNN, whose results and discussions can be found in Appendix \textcolor{blue}{A.3}.

\section{Related Work}

\textbf{Large-batch Optimization.}
For large scale deep model training, it is significant to adopt a larger batch size for better hardware utilization and system scalability. However, large-batch training is prone to converge to a sharp minima, resulting in undesired generalization ability \cite{keskar2016large}. The main reason is that the number of iterations will decrease when we fix the number of epochs in large-batch settings.
Researchers \cite{shallue2019measuring, masters2018revisiting} try to carefully tune the hyper-parameters to narrow this generalization gap. In detail, by incorporating learning rate warm-up and linear scaling,\citet{goyal2017accurate} successfully train ResNet50 with batch size 8192 without loss in generalization performance. Recently, to avoid these hand-tuned methods, the adaptive learning rate on large-batch training has gained enormous attention from researchers.
For example,  LARS and LAMB algorithms {\cite{you2018imagenet,you2019large}} enable researchers to scale the batch size for ResNet50/BERT to 32k/64k.  
Both LARS and LAMB leverage the norm of weights and gradients to adjust the learning rate of each layer. These adaptive methods enable researchers to train ImageNet in a few minutes \cite{jia2018highly,ying2018image,yamazaki2019yet}. {\citet{DBLP:conf/icml/JohnsonAGG20}} propose AdaScale SGD, a novel learning rate schedule rule for stabilizing the warm-up stage. However, it highly depends on the parallelism degree of the system. {\citet{liu2021concurrent}} use adversarial learning to further scale the batch size to 96k. 
More recently, sharpness-aware minimization (SAM) {\cite{foret2020sharpness}} introduces a procedure to minimize the loss value and loss sharpness to close the generalization gap. However, SAM suffers from training efficiency since the update rule of SAM involves two sequential gradient computation at each iteration. There are few works {\cite{liu2022towards, du2021efficient}} towards improving the efficiency of SAM. Recently,  effort {\cite{mccandlish2018empirical}} has been made on how to choose an appropriate batch size and corresponding learning rate for large-batch training.  And {\citet{NEURIPS2021_abea47ba}} propose Simigrad, which utilizes a lightweight and automated adaptive batching method to enable fine-grained adaptive batch size.
However, rather than classification tasks, there are few works towards large-batch training for object detection. {\citet{peng2018megdet}} implement cross-GPU batch normalization to stabilize the training process and {\citet{wang2020large}} propose PMD-LAMB to reduce the negative effects of the lagging historical gradients. They can scale the training of widely used Faster R-CNN+ResNet50 Detector with batch size 256/1056 with small performance drop. 

\textbf{Dense Visual Predictions}
We can divide current deep learning based object detection into two-stage and single-stage detectors. A network that has a separate module to generate region proposals is termed as a two-stage detector. These methods try to find an arbitrary number of proposals in an image during the first stage and then classify and localize them in the second stage, including Faster R-CNN \cite{girshick2015fast}, Mask R-CNN \cite{he2017mask}, and R-FCN \cite{dai2016r}. Single-stage detectors, such as SSD \cite{liu2016ssd} and RetinaNet \cite{lin2017focal}, classify and localize semantic objects in a single shot using dense sampling. They use predefined boxes/keypoints of various scales and aspect ratios to localize objects. Some single-stage detectors, like FOCS \cite{tian2019fcos} can also achieve competitive results with two-stage detectors.
In recent years, deep learning models have yielded a new generation of image segmentation \cite{minaee2021image,ghosh2019understanding} tasks with significant performance improvements. Different from detection tasks, we can group deep learning segmentation based on the segmentation goal into semantic segmentation, instance segmentation, and panoptic segmentation. Semantic segmentation \cite{DBLP:journals/pami/ChenPKMY18, zhao2017pyramid} can be seen as an extension of image classification from image level to pixel level, while instance segmentation \cite{he2017mask,hafiz2020survey} can be defined as the task of finding simultaneous solution to semantic segmentation and object detection. Finally, panoptic segmentation \cite{kirillov2019panoptic,xiong2019upsnet,li2021panoptic} focus on identifying things and stuff separately, also separating (using different colors) the things of the same class.
 
\section{Conclusion}
The complicated pipelines of dense visual predictions suffer from heavy performance drop in large-batch training. In this paper, we propose and fully study AGVM, which enables module-wise learning rate scaling and successfully scales the batch size to larger than 10K with desired generalization performance. We also provide a convergence analysis, showing that AGVM+SGD and AGVM+AdamW both converge to a stable point in the general non-convex setting. Furthermore, we have conducted extensive experiments to show that AGVM can generalize to different complicated pipelines and challenging tasks, including object detection, instance segmentation, semantic segmentation, and panoptic segmentation. We report unprecedented better performance on large-batch training with very large batch size. For example, AGVM trains Faster R-CNN+ResNet50 using batch size of 1536 in 4.2 minutes
without loss of performance. By increasing the object detector UniNet to one billion parameters, AGVM can  achieve 62.2 mAP on COCO using a batch size of 960 in just 3.5 hours, reducing the training time by 20.9$\times$ compared to the normal small-batch training. 

\textbf{Limitation and Potential Negative Societal Impact.}
Module partitioning is important to estimate the effective batch size quantitatively. For some pipelines without explicit modularity such as the heatmap-based pose estimation, we need to do more empirical analysis. We will investigate it in the future. The potential negative social impact is to use the proposed algorithm 
to speed up the training of fraud models such as DeepFake \cite{lyu2020deepfake}.

\begin{ack}
Ping Luo is supported by the General Research Fund of HK No.27208720, No.17212120, and No.17200622.
\end{ack}

\bibliographystyle{unsrt}
\bibliography{egbib}

\appendix
\section{Appendix}

For presenting the details in appendix, we extend the notations as:
given a module set $\mathcal{M}$, e.g., $\mathcal{M}=\{$\texttt{Backbone, FPN, RPN, Detection head}$\}$ for Faster R-CNN, we define $w=\left\{w^{(i)} \mid i \in[1, h]\right\}$ as the weights of it, where $h$ means the number of modules in $\mathcal{M}$ and $w^{(i)}$ indicates the learnable parameters of $i$-th module. Let $w \in \mathbb{R}^{d}$, $w^{(i)} \in \mathbb{R}^{d_{i}}$, and $\Sigma_{i=1}^{h} d_{i}=d$. Given a dataset $S=\left\{\left(x_{i},y_{i}\right)\right\}_{i=1}^{n}$ with $n$ training samples, where $x_{i}$ and $y_{i}$ denote a data point and its label respectively, we can estimate a loss function $L: \mathbb{R}^{d} \rightarrow \mathbb{R}$ for a randomly sampled mini-batch $S_{t}$ to obtain $l\left(w_{t}\right)=\frac{1}{b} \sum_{j \in S_{t}} L\left(w_{t}, (x_{j},y_{j})\right)$, where $S_{t}$ is the mini-batch samples with batch size $\left|S_{t}\right|=b$ at the $t$-th iteration. At the $t$-th backward propagation step, we can derive the gradient $\nabla_{i} l\left(w_{t}\right)$ to update $i$-th module in $\mathcal{M}$. Keep this in mind, we further formulate the gradient of full batch (total samples in $S$) as $\nabla f\left(w_{t}\right)$, where $\nabla f\left(w_{t}\right)=\frac{1}{n} \sum_{j \in S} \nabla L\left(w_{t}, (x_{j},y_{j})\right)$. Naturally, we have $\mathbb{E}\left[\nabla_{i} l\left(w_{t}\right)\right]=\nabla_{i} f\left(w_{t}\right)$. For convenience, we use $g_{t},\|\cdot\|$ and $\|\cdot\|_{1}$ to denote $\nabla l\left(w_{t}\right), l_{2}$-norm and $l_{1}$-norm, respectively. In particular, $g_{t}^{(i)}$ is used to denote $\nabla_{i} l\left(w_{t}\right)$.

\subsection{Gradient Variance Estimation}\label{Appendix:A.1}

\begin{figure}[t] 
\centering 
\includegraphics[width=0.95\textwidth]{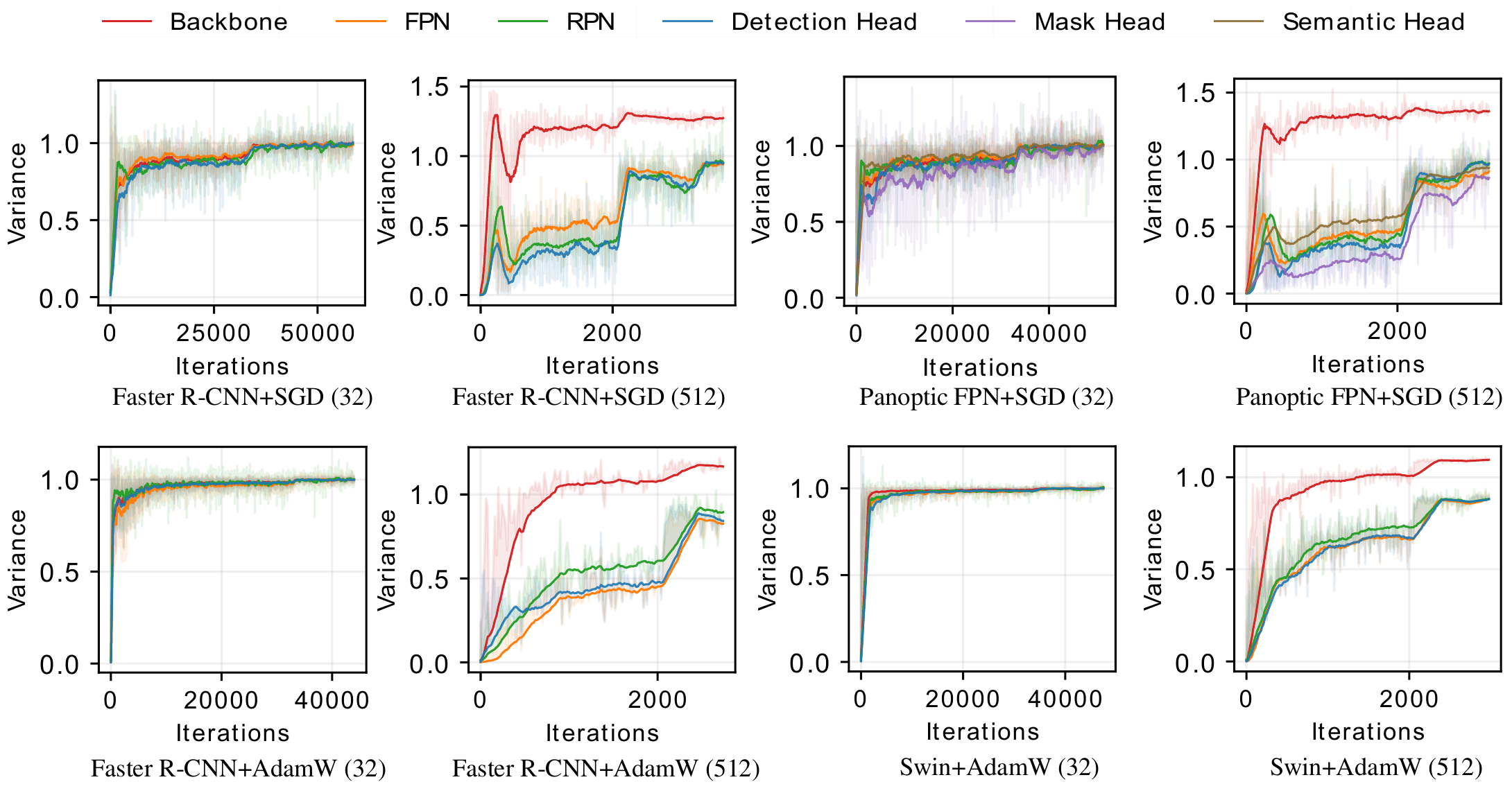}
\caption{\small{\textbf{Comparisons} of the gradient variances (omitting the learning rate $\eta_{t}$ referring to Eq.~\eqref{modified}) in different modules of different pipelines (\ie Faster R-CNN and Panoptic FPN) and optimizers (\ie SGD and AdamW). The number in the bracket represents the batch size. We see that when the batch size is small (\ie 32), the gradient variances are similar. When the batch size is large (\ie 512), the gradient variances all suffer significant misalignment of different modules. All pipelines use ResNet50 as the backbone network other than the last two figures, where we adopt Faster R-CNN+Swin-Tiny to visualize the variances.}}
\label{fig:overall}
\end{figure}
We introduce the gradient variance to measure the gap between SGD (stochastic gradient descent with mini-batch) and GD (gradient descent with full batch). However, computing the accurate gradient variance requires extremely high computational cost and it will slow down training speed dramatically. To address this problem, {\citet{NEURIPS2021_abea47ba}} utilize the cosine similarity between two aggregated gradients from the replicas in a distributed training system to estimate the gradient variance between SGD and GD efficiently.
Specifically, we can compute the gradient for each  sample in the $t$-th  mini-batch $S_t$ of batch size $b$, denoted by $r_{1,t},..., r_{j,t},...,r_{b,t}$, then we have $g_{t} = \frac{1}{b}\sum^b_{j=1}r_{j,t}$. Since we split the above gradients into two groups, averaging each group can obtain $G_{t,1}=\frac{2}{b} \sum_{j=1}^{\frac{b}{2}} r_{2 j-1,t}$ and $G_{t,2}=\frac{2}{b} \sum_{j=1}^{\frac{b}{2}} r_{2 j,t}$, respectively. It formulates
the gradient variance as:
 \begin{equation}
 \mathrm{Var}(g_{t})=\mathbb{E}[\| g_{t}-\nabla f(w_{t})\|^{2}] = \frac{n-b}{2n-b}(1-\mathbb{E}[cos(G_{t,1},G_{t,2})])\mathbb{E}[\| g_{t}\|^{2}],
 \end{equation}
where $n$ and $b$ are the number of training samples and the mini-batch size, respectively. Then we derive a updated (considering learning rate) gradient variance to delve into the difference of network modules in complicated dense visual prediction pipelines. The updated gradient variance of the $i$-th network module at the $t$-th iteration can be formulated as:
\begin{equation}
\mathrm{Var}(\eta_{t} g_t^{(i)})= \mathbb{E}[\|\eta_t g^{(i)}_{t}-\eta_t \nabla_i f(w_{t})\|^{2}]= \frac{n-b}{2n-b}\underbrace{  \eta_{t}^{2}(1-\mathbb{E}[\cos(G_{t,1}^{(i)},G_{t,2}^{(i)})])}_{\Phi^{(i)}_{t}}\mathbb{E}[\| g_t^{(i)}\|^{2}],
\label{modified}
\end{equation}
where $\eta_{t}$ is the learning rate. 
 $G_{t,1}^{(i)}$ and $G_{t,2}^{(i)}$ are two groups of the gradient estimation as discussed above for $i$-th submodule.
Following {\cite{NEURIPS2021_abea47ba,pmlr-v119-wu20c}}, since each entry in the vector $g_t^{(i)}$ could be assumed independent and identically distributed (\textit{i.i.d.}) 
\begin{wrapfigure}[12]{r}{6cm} 
\includegraphics[width=6cm]{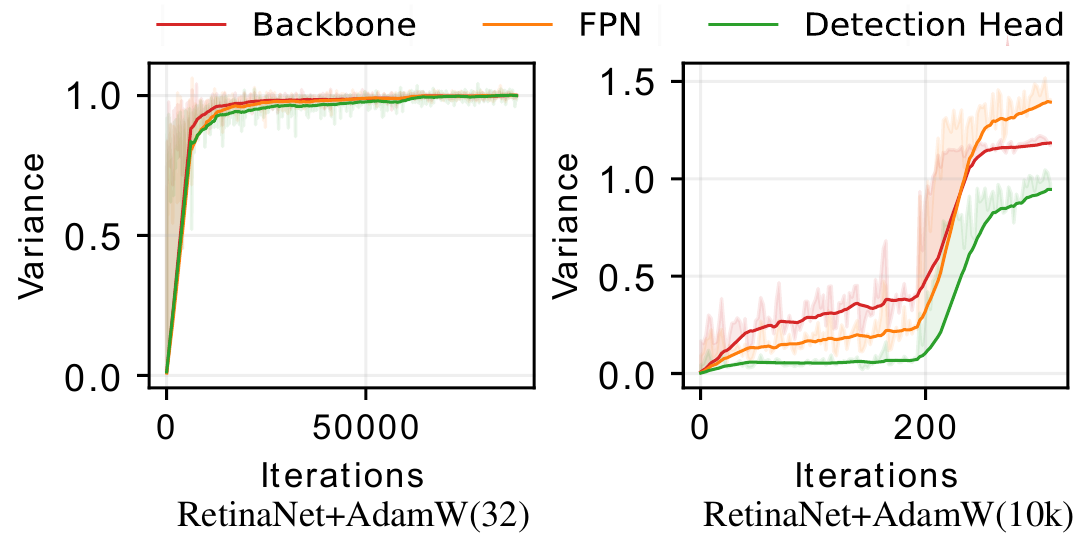}\
\caption{\small{\textbf{Comparisons} of variances for RetinaNet with batch size 32 and 10k}.}
\label{fig:retinanet_4k_10k}
\end{wrapfigure}
in a massive dataset,  $\Phi_t^{(i)}$ is thus proportional to the above updated gradient variance. At each training iteration, we can approximate the updated gradient variance by
$\Phi_t^{(i)} =  \eta_{t}^{2}(1-\cos(G_{t,1}^{(i)},G_{t,2}^{(i)}))$, where $\Phi_t^{(i)}$ indicates the $\mathrm{Var}(\eta_{t} g_t^{(i)})$ normalized by the number of parameters.
For consistency of presentation, we still call $\Phi_t^{(i)}$ gradient variance, which enables us to estimate the gradient variance of each network module at each training iteration. Note that gradient variance magnitude has great influence on the generalization ability of deep neural network \cite{pmlr-v119-wu20c}.
 
\subsection{Overview of Gradient Variance of Different Pipelines}\label{Appendix:A.2}
\begin{figure}[t] 
\centering 
\includegraphics[width=1.0\textwidth]{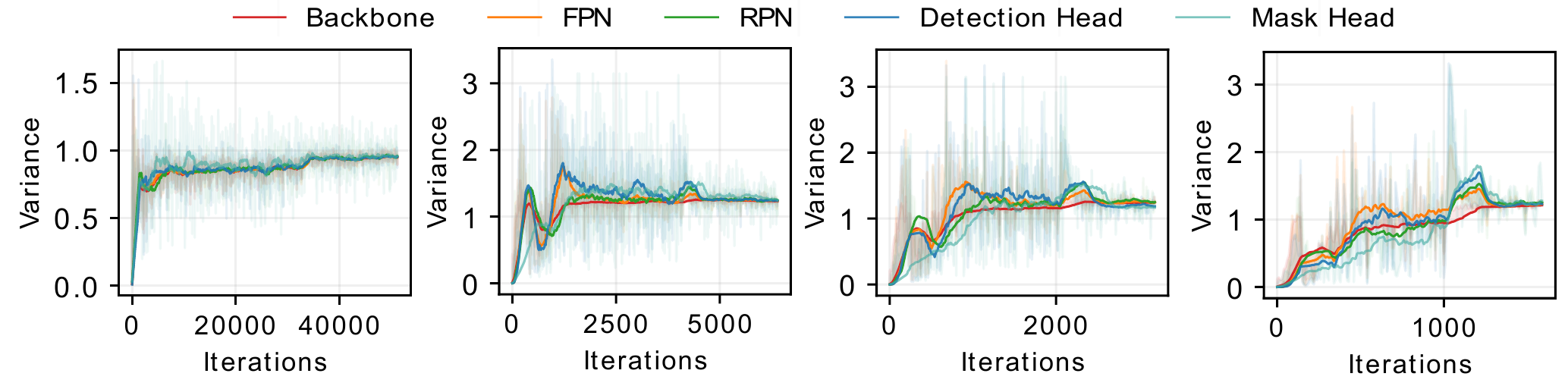}
\caption{\small{\textbf{Comparisons} of the gradient variances of different modules in Mask R-CNN with the help of AGVM. From left to right, the models are trained using SGD with a mini-batch size of 32, 256, 512, and 1024. AGVM helps avoid training failure with batch size 1024.}}
\label{fig:agvm_mask}
\end{figure}

In this section, we give an overview of the gradient variance comparisons of different pipelines in Fig.~\ref{fig:overall}, including four pipelines (\ie Faster R-CNN and Panoptic FPN) and two optimizers (\ie SGD and AdamW). We also show the gradient variances with batch size 32 and 10k in Fig.~\ref{fig:retinanet_4k_10k} on RetinaNet. The variances after applying AGVM on Mask R-CNN is shown in Fig.~\ref{fig:agvm_mask}.


\subsection{Ablation Study of Variance Misalignment on Faster R-CNN}\label{Appendix:A.3}
\begin{figure}[t] 
\centering 
\includegraphics[width=1.0\textwidth]{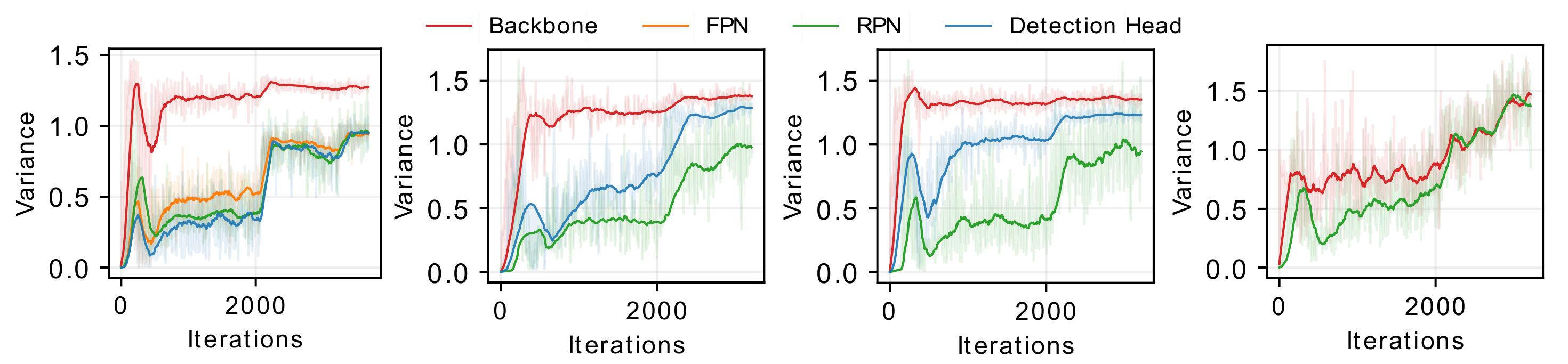}
\caption{\small{
\textbf{Ablative experiments on exploring the gradient variance misalignment.} To validate our result on effective batch size, we progressively remove the FPN, decrease region proposals, and freeze the parameters of detection head to reduce the effective batch size. Finally, it also leads to a variance convergence trend throughout the training between RPN and backbone.}} 
\label{fig:faster_ablation} 
\end{figure}

We define the module set $\mathcal{M}$ as \{\texttt{Backbone}, \texttt{FPN}, \texttt{RPN}, \texttt{Detection head}\} in Faster R-CNN \textcolor{RoyalBlue}{\cite{girshick2015fast}} and $|B_i|$ indicates the \emph{effective batch size} of the $i$-th module in $\mathcal{M}$. Intuitively, there are $|B_4| \approx NK|B_1|$ due to the shared detection head (\ie classifiers/regressors) by all levels of the FPN and different region proposals. $N$ and $K$ indicate the number of FPN levels and region proposals fed into the detection head. To evaluate this assumption, as shown in Fig.~\ref{fig:faster_ablation}, we have three observations. (1) Similar to the ablation study on RetinaNet, we remove the FPN and adopt the final level to perform detection. As illustrated by the second figure in Fig.~\ref{fig:faster_ablation}, the gradient misalignment phenomenon between detection head and backbone has been reduced. (2) Furthermore, we reduce the number of region proposals from 512 to 10. As shown in the third figure in Fig.~\ref{fig:faster_ablation}, this also alleviates the variance difference between detection head and backbone. (3) Finally, we freeze the parameters in the detection head and only train RPN and backbone. Similar to the phenomenon on RetinaNet, this also leads to a variance convergence trend throughout the training between RPN and backbone.
 
\subsection{Proof of Convergence Rate}\label{Appendix:A.4}
In this section, we will show that even using AGVM, SGD and AdamW optimizers still enjoy appealing convergence properties. 
In order to present our analysis, we first need to make some assumptions.

\textbf{Assumptions.} We need to assume function $f(w)$ is $L_{i}- smooth$ with respect to $w^{(i)}$, \ie there exists a constant $L_{i}$ such that: 
\begin{equation}
\forall x, y \in \mathbb{R}^{d},\|\nabla_{i} f(x)-\nabla_{i} f(y)\| \leq L_{i}\|x^{(i)}-y^{(i)}\|,
\end{equation}
for all $i \in[1,h]$. We use $L=\left(L_{1}, \cdots, L_{h}\right)^{\top}$ to denote the $h$-dimensional vector of Lipschitz constants and use $L_{max}$ to denote $\max_{i}L_{i}$. We also assume the following bound on different modules' gradient norm via $\mathbb{E}\left[\|g^{(i)}\|^{2}\right]\leq K\|\nabla_{1} f(w)\|^{2}$. Furthermore, although it's difficult to quantify the effective batch size of different modules, we argue the ratio of effective batch size between different modules should be bounded, so we can assume $1\leq \frac{\mathbb{E}[\|\Phi_t^{(1)}\|]}{\mathbb{E}[\left\|\Phi_t^{(i)}\right\|]} \leq \alpha_{u}$ for $i \in[1,h]$ and $t \in[1,T]$. For the sake of simplicity, we give convergence results when $\beta_{1}=0$ and ignore the weight decay coefficient ($\lambda=0$). However, our analysis should extend to the general case as well.  We leave this investigation in future work.

\subsubsection{Convergence of AGVM+SGD}
For SGD optimizer, we also assume the following bound on the variance in stochastic gradients $\mathbb{E}\left\|g^{(i)}-\nabla_{i} f(w)\right\|^{2} \leq \sigma_{i}^{2}$ for all $w \in \mathbb{R}^{d}$ and $i \in[1,h]$ with effective batch size $b_{i}$.
For component $i$, we have the following update for SGD optimizer:
\begin{equation}
w_{t+1}^{(i)}=w_{t}^{(i)}-\eta_{t}  \sqrt{\frac{\mathbb{E}[\|\Phi_t^{(1)}\|]}{\mathbb{E}[\left\|\Phi_t^{(i)}\right\|]}}g_{t}^{(i)}.
\end{equation}
Since the function $f$ is $L_{i}-smooth$, we can obtain the following inequality:
\begin{equation}
f\left(w_{t+1}\right) \leq f\left(w_{t}\right)+\left\langle\nabla_{i} f\left(w_{t}\right), w_{t+1}^{(i)}-w_{t}^{(i)}\right\rangle+\sum_{i=1}^{h}\eta_{t}^{2} \frac{L_{i}}{2}\frac{\mathbb{E}[\|\Phi_t^{(1)}\|]}{\mathbb{E}[\left\|\Phi_t^{(i)}\right\|]}\left\|g_{t}^{(i)}\right\|^{2}.
\label{eq:sgd_l_smooth}
\end{equation}
Then, we will first give some analysis on the following ratio:
\begin{equation}
\frac{\mathbb{E}[\|\Phi_t^{(1)}\|]}{\mathbb{E}[\left\|\Phi_t^{(i)}\right\|]} = \frac{\mathbb{E}\left[1-cos(G_{t,1}^{(1)},G_{t,2}^{(1)})\right]}{\mathbb{E}\left[1-cos(G_{t,1}^{(i)},G_{t,2}^{(i)})\right]} .
\label{eq:cos_define}
\end{equation}
Because the samples are randomly divided into two groups, according to the law of large numbers, when batch size $b$ goes to infinity, we have:
\begin{equation}
    \mathbb{E}\left[cos(G_{t,1}^{(j)},G_{t,2}^{(j)})\right] \to 1, \forall j\geq 1.
\end{equation}
For $b=2$, each group only has one sample that comes from the same training distribution, we have:
\begin{equation}
    \mathbb{E}\left[cos(G_{t,1}^{(j)},G_{t,2}^{(j)})\right] \to 0, \forall j\geq 1.
\end{equation}
Therefore, there exists a $\hat{b}$ that makes the following equation hold,
\begin{equation}
    \mathbb{E}\left[cos(G_{t,1}^{(j)},G_{t,2}^{(j)})\right] \leq \frac{1}{2}, {\rm if}\, b\leq \hat{b}, \forall j\geq 1.
\end{equation}
Since the effective batch size of backbone is smaller than that of other modules, the gradient variance of backbone is larger than that of other modules, which means:
\begin{equation}
    \mathbb{E}\left[cos(G_{t,1}^{(1)},G_{t,2}^{(1)})\right] \leq \mathbb{E}\left[cos(G_{t,1}^{(i)},G_{t,2}^{(i)})\right], \forall i > 1.
\end{equation}
When $b<\hat{b}$, we further have:
\begin{equation}
    \mathbb{E}\left[cos(G_{t,1}^{(1)},G_{t,2}^{(1)})\right](1 - \mathbb{E}\left[cos(G_{t,1}^{(1)},G_{t,2}^{(1)})\right]) \leq \mathbb{E}\left[cos(G_{t,1}^{(i)},G_{t,2}^{(i)})\right](1 - \mathbb{E}\left[cos(G_{t,1}^{(i)},G_{t,2}^{(i)})\right]), \forall i > 1.
\end{equation}
Based on this, we have the following:
\begin{equation}
\frac{\mathbb{E}\left[1-cos(G_{t,1}^{(1)},G_{t,2}^{(1)})\right]}{\mathbb{E}\left[1-cos(G_{t,1}^{(i)},G_{t,2}^{(i)})\right]}  \leq 
\frac{\mathbb{E}\left[cos(G_{t,1}^{(i)},G_{t,2}^{(i)})\right]}{\mathbb{E}\left[cos(G_{t,1}^{(1)},G_{t,2}^{(1)})\right]} .
\label{eq:cos_eq}
\end{equation}
By displaying $\delta_{t} \equiv g_{t}^{(i)}-\nabla_{i} f\left(w_{t}\right)$, we obtain:  
\begin{equation}
    \mathbb{E}\left[\|g_{t}^{(i)}\|^{2}\right]=\mathbb{E}\left[\|\delta_{t}+ \nabla_{i} f\left(w_{t}\right)\|^{2} \right]\leq \sigma_{i}^{2}+\|\nabla_{i} f\left(w_{t}\right)\|^{2}.
\end{equation}
Following the Eq.(6) in \cite{NEURIPS2021_abea47ba}, we have:
\begin{equation}
\frac{\|\nabla_{i} f(w_{t})\|^{2}}{\|\nabla_{i} f(w_{t})\|^{2}+\sigma_{i}^{2}} \leq  \frac{\|\nabla_{i} f(w_{t})\|^{2}}{\mathbb{E}\left [\|g_{t}^{(i)}\|^{2}\right]}=\mathbb{E}\left[cos(G_{t,1}^{(i)},G_{t,2}^{(i)})\right] \leq 1.
\end{equation}
With the help of above inequality, we have:
\begin{equation}
\frac{\mathbb{E}\left[cos(G_{t,1}^{(i)},G_{t,2}^{(i)})\right]}{\mathbb{E}\left[cos(G_{t,1}^{(1)},G_{t,2}^{(1)})\right]} \leq 1+\frac{\sigma_{1}^{2}}{\|\nabla_{1} f(w_{t})\|^{2}}.
\end{equation}
However, as shown in Fig. \ref{fig:retinanet_4k_10k}, when the batch size is extremely large (\eg 10k), we cannot derive the above inequality. In this case, we have: 
\begin{equation}
\frac{\mathbb{E}\left[1-cos(G_{t,1}^{(1)},G_{t,2}^{(1)})\right]}{\mathbb{E}\left[1-cos(G_{t,1}^{(i)},G_{t,2}^{(i)})\right]} \leq 1+\alpha_{0}+\frac{\sigma_{1}^{2}}{\|\nabla_{1} f(w_{t})\|^{2}},
\label{eq:sgd,cosine}
\end{equation}
where $\alpha_{0}$ is a constant that meets $\alpha_{u}-1-\frac{\sigma_{1}^{2}}{\|\nabla_{1} f(w_{t})\|^{2}} \leq \alpha_{0} \leq \alpha_{u}-1$ for all $t\leq T$. Then by adding Eq.~\eqref{eq:sgd,cosine} to Eq.~\eqref{eq:sgd_l_smooth}, we obtain: 
\begin{equation}
    f\left(w_{t+1}\right) \leq f\left(w_{t}\right)+\left\langle\nabla_{i} f\left(w_{t}\right), w_{t+1}^{(i)}-w_{t}^{(i)}\right\rangle + \sum_{i=1}^{h}\eta_{t}^{2} \frac{L_{i}}{2}\left(\alpha_0+1+\frac{\sigma_{1}^{2}}{\|\nabla_{1} f(w_{t})\|^{2}}\right)\|g_{t}^{(i)}\|^{2}.
\end{equation}
Taking expectation on the both side, according to the assumption on Eq.~\eqref{eq:cos_define}, we have:
\begin{equation}
\begin{split}
\mathbb{E}\left[f\left(w_{t+1}\right)\right] & \leq  f\left(w_{t}\right)- \eta_{t} \sum_{i=1}^{h}\|\nabla_{i} f\left(w_{t}\right)\|^{2}+\sum_{i=1}^{h}\eta_{t}^{2} \frac{L_{i}}{2}\left(\alpha_{0}+1+\frac{\sigma_{1}^{2}}{\|\nabla_{1} f(w_{t})\|^{2}}\right)\mathbb{E}\left[\|g_{t}^{(i)}\|^{2}\right]
\\& \leq  f\left(w_{t}\right)- \eta_{t} \sum_{i=1}^{h}\|\nabla_{i} f\left(w_{t}\right)\|^{2}+\sum_{i=1}^{h}\eta_{t}^{2} \frac{L_{i}}{2}\left((1+\alpha_{0})\mathbb{E}\left[\|g_{t}^{(i)}\|^{2}\right]+K\sigma_{1}^{2}\right)
\\&\leq  f\left(w_{t}\right)- \eta_{t} \sum_{i=1}^{h}\|\nabla_{i} f\left(w_{t}\right)\|^{2}+\sum_{i=1}^{h}\eta_{t}^{2} \frac{L_{i}}{2}\left((1+\alpha_{0})(\sigma_{i}^{2}+\|\nabla_{i} f\left(w_{t}\right)\|^{2})+K\sigma_{1}^{2}\right)
\\& =  f\left(w_{t}\right)-  \sum_{i=1}^{h}\left(\eta_{t}-\frac{L_{max}}{2}(1+\alpha_0)\eta^{2}_{t}\right)\|\nabla_{i} f\left(w_{t}\right)\|^{2}+\sum_{i=1}^{h}\eta_{t}^{2} \frac{L_{i}}{2}\left(K\sigma_{1}^{2}+(1+\alpha_{0})\sigma_{i}^{2}\right).
\end{split}
\end{equation}
Summing both sides of this inequality and taking the complete expectation, we get:
\begin{equation}
\begin{split}
 \mathbb{E}\left[f\left(w_{t+1}\right)\right] &\leq  f\left(w_{1}\right)\\&-\sum_{t=1}^{T}\sum_{i=1}^{h}\left(\eta_{t}-\frac{L_{max}}{2}\eta^{2}_{t}(1+\alpha_0)\right)\mathbb{E}[\|\nabla_{i} f\left(w_{t}\right)\|^{2}]+T\sum_{i=1}^{h}\eta_{t}^{2} \frac{L_{i}}{2}\left(K\sigma_{1}^{2}+(1+\alpha_{0})\sigma_{i}^{2}\right)  .
\end{split}
\end{equation}
Define $f_{inf} = \inf f\left(w_{t} \right)$ and arrange the above inequality, we can get:
\begin{equation}
\frac{1}{T}\sum_{t=1}^{T}\sum_{i=1}^{h}\mathbb{E}\left[ \|\nabla_{i} f\left(w_{t}\right)\|^{2}\right]\leq \frac{f\left(w_{1}\right)-f_{inf}}{T\left(\eta_{t}-\frac{L_{max}}{2}\eta^{2}_{t}(1+\alpha_{0})\right)}+\frac{\sum_{i=1}^{h}\eta_{t} L_{i}\left(K\sigma_{1}^{2}+(1+\alpha_0)\sigma_{i}^{2}\right)   }{2-L_{max}\eta_{t}(1+\alpha_0)}.
\end{equation}
Let $\eta_{t}\leq \frac{1}{(1+\alpha_0)L_{max}}$, we have the following bound:
\begin{equation}
\frac{1}{T}\sum_{t=1}^{T}\mathbb{E}\left[ \|\nabla f\left(w_{t}\right)\|^{2}\right]\leq \frac{2\left(f\left(w_{1}\right)-f_{inf}\right)}{T\eta_{t}}+\sum_{i=1}^{h}\eta_{t} L_{i}\left(K\sigma_{1}^{2}+(1+\alpha_{0})\sigma_{i}^{2}\right).   
\label{eq:sgd,final}
\end{equation}
\subsubsection{Convergence of AGVM+AdamW}
For AdamW optimizer, we also assume $\|g_{t}\|_{\infty} \leq G-\sqrt{\epsilon}$, $d_{i}=\frac{d}{h}$.
Following \cite{defossez2020simple}, we rewrite the learning rate in the following manner: $\tilde{\eta_{t}}=\eta_{t}\sqrt{\frac{1-\beta_{2}^{t}}{1-\beta_{2}}}$. Based on this, we can redefine the $v_{t}$ as $v_{t} = \beta_{2} v_{t-1} +  g_{t}^{2}$, and let $\tilde{v}_{t}=\beta_{2} v_{t-1} +  \mathbb{E}[g_{t}^{2}]$. So the update of original AdamW can be given by:$r_{t}=\frac{g_{t}}{\sqrt{v_{t}+\epsilon}}$, then we have the following update for AGVM+AdamW:  
\begin{equation}
w_{t+1}^{(i)}=w_{t}^{(i)}-\tilde{\eta}_{t}  \sqrt{\frac{\mathbb{E}[\|\Phi_t^{(1)}\|]}{\mathbb{E}[\left\|\Phi_t^{(i)}\right\|]}}r_{t}^{(i)}.
\end{equation}
Since the function $f$ is $L_{i}-smooth$, we have the following:
\begin{equation}
f\left(w_{t+1}\right) \leq f\left(w_{t}\right)+\left\langle\nabla_{i} f\left(w_{t}\right), w_{t+1}^{(i)}-w_{t}^{(i)}\right\rangle+\sum_{i=1}^{h}\tilde{\eta}_{t}^{2} \frac{L_{i}}{2}\frac{\mathbb{E}[\|\Phi_t^{(1)}\|]}{\mathbb{E}[\left\|\Phi_t^{(i)}\right\|]}\left\|r_{t}^{(i)}\right\|^{2}.
\label{eq:adam,l-smooth}
\end{equation}
For any component $i$, we have:
\begin{equation}
\mathbb{E}\left[cos(G_{t,1}^{(i)},G_{t,2}^{(i)})\right]=\frac{\sum_{j=0}^{d_{i}}(\mathbb{E}\left[g_{t,j}^{(i)}/\sqrt{\epsilon+v_{t,j}^{(i)}}\right])^{2}}{\sum_{j=0}^{d_{i}}\mathbb{E}\left [(g_{t,j}^{(i)}/\sqrt{\epsilon+v_{t,j}^{(i)}})^{2}\right]} \leq 1,
\end{equation}
where $g_{t,j}^{(i)}$ and $v_{t,j}^{(i)}$ denote the $j$-th entry of $g_{t}^{(i)}$ and $v_{t}^{(i)}$. Thanks to the $l_{\infty}$ bound on $g_{t}$, we have $g_{t}^{(i)} \leq \sqrt{\epsilon+v_{t,j}^{(i)}} \leq \frac{G}{\sqrt{1-\beta_{2}}}$, so that:
\begin{equation}
\frac{\|\nabla_{i} f(w_{t})\|^{2}}{d_{i}(G^{2}/(1-\beta_{2}))} \leq \frac{\sum_{j=0}^{d_{i}}(\mathbb{E}\left[g^{(i)}_{t}/\sqrt{\epsilon+v_{t,j}^{(i)}}\right])^{2}}{\sum_{j=0}^{d_{i}}\mathbb{E}\left [(g_{t}^{(i)}/\sqrt{\epsilon+v_{t,j}^{(i)}})^{2}\right]}   \leq 1.
\end{equation}
Similar to Eq.~\eqref{eq:cos_eq}, then we get:
\begin{equation}
\frac{\mathbb{E}[\|\Phi_t^{(1)}\|]}{\mathbb{E}[\left\|\Phi_t^{(i)}\right\|]} = 
\frac{\mathbb{E}\left[1-cos(G_{t,1}^{(1)},G_{t,2}^{(1)})\right]}{\mathbb{E}\left[1-cos(G_{t,1}^{(i)},G_{t,2}^{(i)})\right]}  \leq    \frac{\mathbb{E}\left[cos(G_{t,1}^{(i)},G_{t,2}^{(i)})\right]}{\mathbb{E}\left[cos(G_{t,1}^{(1)},G_{t,2}^{(1)})\right]} \leq \frac{d_{1}(G^{2}/(1-\beta_{2}))}{\|\nabla_{1} f(w_{t})\|^{2}}.
\label{eq:to bound}
\end{equation}
However, since $1-\beta_{2} \to 0$ in general AdamW settings, 
as well as for some extremely large batch size settings (where the upper bound of Eq.~\eqref{eq:to bound} is dominated by $\alpha_{u}$), we have the following for the sake of consistency:
\begin{equation}
    \frac{\mathbb{E}\left[1-cos(G_{t,1}^{(1)},G_{t,2}^{(1)})\right]}{\mathbb{E}\left[1-cos(G_{t,1}^{(i)},G_{t,2}^{(i)})\right]}  \leq \min\{\frac{d_{1}(G^{2}/(1-\beta_{2}))}{\|\nabla_{1} f(w_{t})\|^{2}}, \alpha_{u}\}.
    \label{eq:two item}
\end{equation}
We will give the convergence bounds using these two items, respectively. For the first item, 
by rewriting Lemma 1 in \cite{defossez2020simple}, we get:
\begin{equation}
\mathbb{E}{\left[\nabla_{i,j} f\left(w_{t}\right) \frac{g_{t,j}^{(i)}}{\sqrt{\epsilon+v_{t, j}^{(i)}}}\right] \geq \frac{\left(\nabla_{i,j} f\left(w_{t}\right)\right)^{2}}{2 \sqrt{\epsilon+\tilde{v}_{t, j}^{(i)}}} } 
-2 G \mathbb{E}\left[\frac{\left(g_{t,j}^{(i)}\right)^{2}}{\epsilon+v_{t, j}^{(i)}}\right],
\end{equation}
where we denote the $j$-th entry of $\nabla_{i}f(w_{t})$ by $\nabla_{i,j}f(w_{t})$. Thanks to the $l_{\infty}$ bounded on $g^{(i)}$, we have:
\begin{equation}
\tilde{\eta}_{t} \frac{\left(\nabla_{i,j} f\left(w_{t}\right)\right)^{2}}{2 \sqrt{\epsilon+\tilde{v}_{t,j}^{(i)}}} \geq \frac{\eta_{t}\left(\nabla_{i,j} f\left(w_{t}\right)\right)^{2}}{2 G}.
\label{eq:lemma1}
\end{equation}
Taking expectation on Eq.~\eqref{eq:adam,l-smooth}, and adding Eq.~\eqref{eq:lemma1} to Eq.~\eqref{eq:adam,l-smooth}, we have:
\begin{equation}
\begin{split}
\mathbb{E}\left[f\left(w_{t+1}\right)\right]  & \leq f\left(w_{t}\right) \\& -\frac{\eta_{t}}{2 G}\|\nabla f\left(w_{t}\right)\|^{2}
 +\sum_{i=1}^{h}\left(2 \tilde{\eta}_{t} G+\frac{\tilde{\eta}_{t}^{2} L_{i}d_{1}(G^{2}/(1-\beta_{2}))}{2\|\nabla_{1} f(w_{t})\|^{2}}\right) \mathbb{E}\left[\left\|r_{t}^{(i)}\right\|^{2}\right].
\label{eq:adamw:before sum up}
\end{split}
\end{equation}
Taking complete expectation on Eq.~\eqref{eq:adamw:before sum up} and sum up:
\begin{equation}
\begin{split}
    \mathbb{E}\left[f\left(w_{t+1}\right)\right]  & \leq f\left(w_{1}\right)  -\frac{\eta_{t}}{2 G}\sum_{t=1}^{T}\mathbb{E}\left[ \|\nabla f\left(w_{t}\right)\|^{2}\right]
 \\&+\sum_{t=1}^{T}\sum_{i=1}^{h}\left(\frac{2 \eta_{t} G}{\sqrt{1-\beta_{2}}}\mathbb{E}\left[\left\|r_{t}^{(i)}\right\|^{2}\right]\right)+\frac{\eta_{t}^{2} \|L\|_{1}d_{1}(G^{2}/(1-\beta_{2}))KT}{2\epsilon(1-\beta_{2})} .
 \end{split}
\end{equation}
Then, with the help of Lemma 2 in \cite{defossez2020simple}, we get:
\begin{equation}
\begin{split}
    \mathbb{E}\left[f\left(w_{t+1}\right)\right]  & \leq f\left(w_{1}\right)  -\frac{\eta_{t}}{2 G}\sum_{t=1}^{T}\mathbb{E}\left[ \|\nabla f\left(w_{t}\right)\|^{2}\right]
 \\&+\frac{2 \eta_{t} Gd}{\sqrt{1-\beta_{2}}}\left(\frac{1}{T}\ln\left(1+\frac{G^{2}}{(1-\beta_{2})\epsilon}\right)-T\ln(\beta_{2})\right)+\frac{\eta_{t}^{2} \|L\|_{1}d_{1}(G^{2}/(1-\beta_{2}))KT}{2\epsilon(1-\beta_{2})}.
 \end{split}
\end{equation}
For the second item in Eq.~\eqref{eq:two item}, taking complete expectation on Eq.~\eqref{eq:adam,l-smooth} and sum up:
\begin{equation}
\begin{split}
    \mathbb{E}\left[f\left(w_{t+1}\right)\right] &  \leq f\left(w_{1}\right) \\& -\frac{\eta_{t}}{2 G}\sum_{t=1}^{T}\mathbb{E}\left[ \|\nabla f\left(w_{t}\right)\|^{2}\right]
     +\sum_{t=1}^{T}\sum_{i=1}^{h}\left(\left(\frac{2 \eta_{t} G}{\sqrt{1-\beta_{2}}}+\tilde{\eta}_{t}^{2} \alpha_{u} \frac{L_{i}}{2}\right)\mathbb{E}\left[\left\|r_{t}^{(i)}\right\|^{2}\right]\right).
 \end{split}
\end{equation}
With the help of Lemma 2 in \cite{defossez2020simple}, we get:
\begin{equation}
\begin{split}
    \mathbb{E}\left[f\left(w_{t+1}\right)\right] &  \leq f\left(w_{1}\right) -\frac{\eta_{t}}{2 G}\sum_{t=1}^{T}\mathbb{E}\left[ \|\nabla f\left(w_{t}\right)\|^{2}\right]
   \\&   +\left(\frac{2 \eta_{t} Gd}{\sqrt{1-\beta_{2}}}+\tilde{\eta}_{t}^{2}\alpha_{u}h \frac{\|L\|_{1}}{2}\right)\left(\frac{1}{T}\ln\left(1+\frac{G^{2}}{(1-\beta_{2})\epsilon}\right)-T\ln(\beta_{2})\right).
 \end{split}
\end{equation}
Finally, we have:
\begin{equation}
\begin{split}
&\frac{1}{2 GT}\sum_{t=1}^{T}\mathbb{E}\left[ \|\nabla f\left(w_{t}\right)\|^{2}\right]  \leq \frac{f\left(w_{1}\right)-f_{inf}}{\eta_{t}T}+ \frac{2  G d}{\sqrt{1-\beta_{2}}}\left(\frac{1}{T}\ln\left(1+\frac{G^{2}}{(1-\beta_{2})\epsilon}\right)-\ln(\beta_{2})\right) +C, \\&
C=\min\left\{\frac{\eta_{t} \|L\|_{1}dG^{2}K}{2\epsilon h(1-\beta_{2})^{2}}, \frac{\eta_{t}\alpha_{u}h\|L\|_{1}}{2(1-\beta_{2})}\left(\frac{1}{T}\ln\left(1+\frac{G^{2}}{(1-\beta_{2})\epsilon}\right)-\ln(\beta_{2})\right)\right\}.
 \end{split}
 \label{eq:adamw_final}
\end{equation}
For AGVM+SGD, suppose $\eta_{t}=\frac{1}{\sqrt{T}}$, and for AGVM+AdamW, let $\eta_{t}=\frac{1}{\sqrt{T}}$ and $\beta_{2}=1-\frac{1}{T}$, then SGD and AdamW achieve $O(1/\sqrt{T})$ and $O(\ln(T)/\sqrt{T})$ convergence rate, respectively. Note that in this case, the upper bound of Eq.~\eqref{eq:adamw_final} is dominated by the second item of $C$. 
\subsubsection{Linear Speedup Property of AGVM}
We give the linear speedup property for AGVM+synchronous SGD w.r.t. batch size as a corollary.  First, we will prove gradient variance decreases linearly with batch size $b$.  
For ease of understanding, we assume that $\nabla f(w)$, $g$, $r$ represent the gradient of the full dataset, the mini-batch with size $b$ and the single sample, respectively. Then we have the following covariance matrix:
\begin{equation}
    \Sigma(w):=\operatorname{cov}\left[r\right]=
\frac{1}{n} \sum_{i=1}^{n}\left(r_i-\nabla f(w)\right)\left(r_i-\nabla f(w)\right)^{T},
\end{equation}
where $n$ indicates the total number of training samples. Likewise, a stochastic gradient $g$ computed on a randomly-drawn mini-batch is a random variable with mean $\nabla f(w)$. Assuming that it is composed of $b$ samples drawn independently with replacement, its covariance matrix is:
\begin{equation}
    \operatorname{cov}[g]=\frac{\Sigma(w)}{b}.
\end{equation}
According to the Central Limit Theorem, g can be approximately normally distributed:
\begin{equation}
    g \sim \mathcal{N}\left(\nabla f(w), \frac{\Sigma(w)}{b}\right).
\end{equation}
As assumed in Appendix A.4.1 section, the variance of stochastic gradients with batch size $b_i$ meets $\mathbb{E}\left\|g^{(i)}-\nabla_{i} f(w)\right\|^{2} \leq \sigma_{i}^{2}$ for all $w \in \mathbb{R}^{d}$ and $i \in[1,h]$. 
So when we increase the batch size from $b_i$ to $Mb_i$, we have:
\begin{equation}
    \mathbb{E}\left\|g^{(i)}-\nabla_{i} f(w)\right\|^{2} \leq \frac{\sigma_{i}^{2}}{M}.
\end{equation}
By substituting $\sigma_{i}^{2}$ with $ \frac{\sigma_{i}^{2}}{M}$ for all $i \in [1,h]$ in Eq.\eqref{eq:sgd,final}, we get:
\begin{equation}
    \frac{1}{T}\sum_{t=1}^{T}\mathbb{E}\left[ \|\nabla f\left(w_{t}\right)\|^{2}\right]\leq \frac{2\left(f\left(w_{1}\right)-f_{inf}\right)}{T\eta_{t}}+\sum_{i=1}^{h}\eta_{t} L_{i}\left(K\frac{\sigma_{1}^{2}}{M}+(1+\alpha_{0})\frac{\sigma_{i}^{2}}{M}\right).
\end{equation}
Let $\eta_{t}=\sqrt{\frac{M}{T}}$, we obtain a $O(1/\sqrt{MT})$ convergence rate.

\subsection{Parameter Settings}\label{Appendix:A.5}
\subsubsection{Settings for Different Visual Predictors}
In this section, we give the detailed hyper-parameter settings for the training of different visual predictors, which are shown in Table~\ref{tab:setting_sgd}, Table~\ref{tab:setting_sgd_sem} and Table~\ref{tab:setting_adamw}.
All predictors are evaluated on the \textbf{validation set} of COCO and ADE20K datasets.
For SGD optimizer, we do not follow the linear learning rate scaling in \cite{wang2020large} since the large learning rate on batch size 512 leads to the training failure of baseline.
Instead, when the batch size is greater than 128 (256 for semantic segmentation), we use the square root of learning rate scaling to avoid divergence in the training process. 
With this strategy, we obtain a better baseline than \cite{wang2020large}. Especially, the best learning rate on Faster R-CNN on batch size 512 is 0.38.
For AdamW optimizer, the learning rate scaling strategy is almost the same as SGD. The only difference is that we adopt a smoother scaling scheme due to its faster convergence speed.
Specifically, when the batch size is greater than 128, the learning rate is scaled up with a ratio of $\sqrt{1.5}$ if we double the batch size. 

\begin{table}[H]
\centering
\caption{\small{Hyper-parameter settings for SGD optimizer on Faster R-CNN, Mask R-CNN, and Panoptic FPN with the CNN backbone. LR represents the global learning rate.}}
\scalebox{0.9}{
\begin{tabular}{ccccccc} 
\toprule
 Batch Size &  Warmup Epochs & LR & LR Decay &  $\tau$ & $\alpha$ & Weight Decay  \\
\midrule  32  & 1 & 0.04 & MultiStep & 10 & 0.97 & 1e-4  \\
 256  & 2 & 0.226 & MultiStep & 10 & 0.97 & 1e-4  \\
 512  & 2 & 0.32 & MultiStep & 10 & 0.97 & 1e-4  \\
 1024  & 2 & 0.452 & MultiStep & 5 & 0.97 & 1e-4 \\
\bottomrule
\end{tabular}}
\label{tab:setting_sgd}
\end{table}

\begin{table}[H]
\centering
\caption{\small{Hyper-parameter settings for SGD optimizer on Semantic FPN with the CNN backbone. LR represents the global learning rate. "Poly" means that the learning rate at current iteration is multiplied by $(1 - \frac{iter}{max\_ iter})^{power}$ (with $power=0.9$).}}
\scalebox{0.9}{
\begin{tabular}{ccccccc} 
\toprule
 Batch Size &  Warmup Iters & LR & LR Decay &  $\tau$ & $\alpha$ & Weight Decay  \\
\midrule  
 32   & 500 & 0.01 & Poly & 5 & 0.97 & 5e-4  \\
 512  & 500 & 0.113 & Poly & 5 & 0.97 & 5e-4  \\
 1024 & 250 & 0.16 & Poly & 5 & 0.97 & 5e-4  \\
 2048 & 125 & 0.226 & Poly & 5 & 0.97 & 5e-4 \\
\bottomrule
\end{tabular}}
\label{tab:setting_sgd_sem}
\end{table}

\begin{table}[H]
\centering
\caption{\small{Hyper-parameter settings for AdamW optimizer on Faster R-CNN with the Tranformer backbone. LR represents the global learning rate.}}
\scalebox{0.9}{
\begin{tabular}{cccccccc}
\toprule
 Batch Size  & Warmup Epochs & LR & LR Decay &  $\tau$ & $\alpha$ & Weight Decay  &Gradient Clip \\
\midrule 
 32 &  1 & 2e-4 & MultiStep & 10 & 0.97 & 0.05 &- \\
 256 &  2 & 9.8e-4 & MultiStep & 10 & 0.97 & 0.05 & 1.0 \\
 512 & 2 & 1.2e-3 & MultiStep & 10 & 0.97 & 0.05 &1.0 \\
 1024 &  3 & 1.5e-3 & MultiStep & 5 & 0.97 & 0.05 &1.0 \\
\bottomrule
\end{tabular}}
\label{tab:setting_adamw}
\end{table}

\subsubsection{Settings for Billion-level UniNet}
\begin{table}[H]
    \centering
    \caption{UniNet-G architecture. We adopt the Fused MBConv blocks \cite{tan2021efficientnetv2} and transformer blocks to form a hybrid convolution-transformer visual network.}
    \scalebox{0.87}{
    \begin{tabular}{cccccc}
    \toprule
    \multirow{2}{*}{Stage}  & \multirow{2}{*}{Block} & \multicolumn{4}{c}{Network Size} \\
     &  & Expansion & Channel & Layers & Stride \\
    \midrule
    0     & Fused MBConv & 1 & 104 & 6 & 2 \\
    1     & Fused MBConv & 4 & 216 & 9 & 4 \\
    2     & Fused MBConv & 6 & 384 & 18 & 8 \\
    3     & Fused MBConv & 3 & 576 & 18 & 16 \\
    4     & Transformer & 2 & 576 & 18 & 16 \\
    5     & Transformer & 5 & 1152 & 36 & 32  \\
    \bottomrule
    \end{tabular}}
    \label{tab:uninet_b05}
\end{table}

We scale the UniNet \cite{liu2021uninet} to 1-billion parameters and evaluate it on COCO \textbf{test-dev} benchmark.
The detailed architecture is presented in Table \ref{tab:uninet_b05}. 

\paragraph{Improved HTC detector.}
To compare with the state-of-the-art, we implement some extensions to the original HTC \cite{htc} and denote it as HTC-X. 
This improved version is built upon the light-weight variant of HTC (HTC-Lite \cite{htc_lite}).
To reduce the computation overheads, the transformer blocks of UniNet-G backbone are evenly split into 18 subsets. 
There are two blocks using window attention and the last block using global attention in each subset.
Furthermore, we adopt RCNet \cite{rcnet} and SEPC \cite{sepc} as the feature pyramid with levels from $P_{3}$ to $P_{8}$, and increase the feature channel from 256 to 384.
The positive IoU thresholds in the R-CNN stage are increased to 0.6, 0.7, 0.8.
We use 4 decoupled transformer blocks for the classification branch and localization branch, respectively.

\paragraph{ImageNet-22K pre-training.} 
We train the UniNet-G for 150 epochs using an AdamW optimizer and a cosine learning rate scheduler.
The peak learning rate is $0.005$ and the minimum learning rate is $0.0001$.
A batch size of 5120 and a weight decay coefficient of 0.03 are used.
We adopt common augmentation techniques including Mixup, Cutmix, Random Erasing, and stochastic depth with a ratio of 0.3.

\paragraph{Finetuning on COCO object detection.}
We first finetune the improved HTC-X (without the mask branch) on the Objects-365 V1 dataset \cite{shao2019objects365}, which consists of 638k images. 
The model is trained with an AdamW optimizer with a learning rate of $8e-5$ and a batch size of 64 for 20 epochs.
Then we further finetune it on COCO dataset for only 11 epochs.
A batch size of 960 and a learning rate of $1.5e-4$ are adopted.
During the finetuning phase, the shorter side of the input image is randomly selected between 400 and 1200 while the longer side is at most 1600.
The window sizes of UniNet-G are set to $28\times28$ for Stage 4 and $14\times14$ for Stage 5.

\subsection{Overview of AGVM-enabled SGD and AdamW}\label{Appendix:overview}
 We treat  the $\texttt{Backbone}$ ($i=1$)  as the anchor and modulate other modules making their gradient variances consistent with the $\texttt{Backbone}$.
Specifically, we adjust the module learning rates $\hat{\eta}_t^{(i)}$ by using
the ratio between $\Phi^{(1)}_t$ and $\Phi^{(i)}_t$.
The update rule for each network module can be written as:
\begin{equation}
w^{(i)}_{t+1} = w^{(i)}_{t}-\hat{\eta}_t^{(i)} g^{(i)}_{t},~~\mathrm{where~~} \hat{\eta}_t^{(i)} = \eta_{t} \mu^{(i)}_t ~~\mathrm{and~~} \mu^{(i)}_t=\sqrt{\frac{\Phi^{(1)}_t }{\Phi^{(i)}_t}},
\label{sgd-gvc}
\end{equation}
where $ \eta_{t}$ is the global learning rate. 
However, simply adjusting the learning rates on-the-fly would easily yield training failure due to the transitory large variance ratio that impedes the optimization. We propose a momentum update to address this problem.
Let $\alpha\in[0,1)$ be a momentum coefficient, we have:
\begin{equation}
\mu_t^{(i)} \gets \alpha \mu_{t-1}^{(i)} + (1-\alpha)\mu_t^{(i)},
\label{moving}
\end{equation}
which can reduce the influence of unstable variance. Note that we update $\mu_t^{(i)}$ each $\tau$ iterations. Based on this, we present AGVM-enabled SGD and AdamW optimizers in Alg.~\ref{alg:SGD}, and Alg.~\ref{alg:lamb}. In the practical implementation in extremely-large batch regime (\eg 10k), we add a small epsilon value $\mu_{t}^{(i)}=\sqrt{\frac{\Phi_{t}^{(1)}+\epsilon}{\Phi_{t}^{(i)}+\epsilon}}$ in Eq.(41) to ensure stability and also clip the $\mu_{t}^{(i)}$ to [0.1, 10].
\begin{figure}
\begin{minipage}[b]{.48\textwidth}
\begin{algorithm}[H]\small
	\caption{AGVM+SGD}
	\label{alg:SGD}
	\begin{algorithmic}
		\STATE {\bfseries Input:} $w_1 \in \mathbb{R}^d$, learning rate $\{\eta_t\}_{t=1}^T$, parameters $0 \leq \beta_{1},\alpha < 1$, interval $\tau$, weight decay coefficient $\lambda$
		\STATE Set $m_{0} = 0$, $u_{0}^{(i)}=1$ for $i \in [1,h]$
		\FOR{$t=1$ {\bfseries to} $T$}
		\STATE Draw b samples $S_t$ from dataset $S$
        \STATE Compute $g_{t} = \frac{1}{b} \sum_{j \in S_{t}} \nabla
        l\left(w_{t},(x_{j},y_{j})\right)$
        \IF{\textcolor{blue}{$t\% \tau = 0$}}
        \STATE \textcolor{blue}{Compute $\Phi^{(i)}_{t}$ via gradients $g^{(i)}_{t}$}
        \STATE\textcolor{blue}{ Compute $\hat{\eta_{t}}^{(i)}$ and $\mu_{t}^{(i)}$}
		\ENDIF
        \STATE $m_{t} = \beta_{1} m_{t-1} + (1 - \beta_{1}) (g_{t}+\lambda w_t)$
		\STATE \textcolor{blue}{$w_{t+1}^{(i)} = w_{t}^{(i)} - \hat{\eta_{t}}^{(i)} m_t^{(i)}$ }
		\ENDFOR
	\end{algorithmic}
\end{algorithm}
\end{minipage}\hfill
\begin{minipage}[b]{.5\textwidth}
\begin{algorithm}[H]\small
	\caption{AGVM+AdamW}
	\label{alg:lamb}
	\begin{algorithmic}
		\STATE {\bf Input:} $w_1 \in \mathbb{R}^d$, learning rate $\{\eta_t\}_{t=1}^T$,  parameters $0 \leq \beta_{1}, \beta_2 ,\alpha< 1$, interval $\tau$, weight decay coefficient $\lambda$
		\STATE Set $m_{0} = 0$, $v_{0} = 0$,  $u_{0}^{(i)}=1$ for $i \in [1,h]$
		\FOR{$t=1$ {\bf to} $T$}
		\STATE Draw b samples $S_t$ from dataset $S$
        \STATE Compute $g_{t} = \frac{1}{b} \sum_{j \in S_{t}} \nabla
        l\left(w_{t},(x_{j},y_{j})\right)$
		\STATE  $m_{t} = \beta_{1} m_{t-1} + (1 - \beta_{1}) g_{t}$ 
		\STATE  $v_{t} = \beta_{2} v_{t-1} + (1 - \beta_{2}) g_{t}^2$
		\IF{ \textcolor{blue}{$t\% \tau = 0$}}
        \STATE \textcolor{blue}{Compute $\Phi^{(i)}_{t}$ via modified gradients $\frac{g^{(i)}_{t}}{ \sqrt{v_{t}+\epsilon}}$}
        \STATE  \textcolor{blue}{Compute $\hat{\eta_{t}}^{(i)}$ and $\mu_{t}^{(i)}$}
		\ENDIF
		\STATE $m_t =\frac{ m_t}{1 - {\beta}_1^t}$,
         $v_t = \frac{v_t}{1 - {\beta}_2^t}$,
         $r_{t}=\frac{m_{t}}{\sqrt{v_{t}+\epsilon}}$
		\STATE\textcolor{blue}{ $w_{t+1}^{(i)} = w_{t}^{(i)} - \hat{\eta_{t}}^{(i)} (r_{t}^{(i)} + \lambda w_t^{(i)})$}
		\ENDFOR
	\end{algorithmic}
\end{algorithm}
\end{minipage}
\end{figure}

\end{document}